\documentclass[letterpaper]{article} 
\usepackage{aaai24}  
\usepackage{times}  
\usepackage{helvet}  
\usepackage{courier}  
\usepackage[hyphens]{url}  
\usepackage{graphicx} 
\usepackage{natbib}  
\usepackage{caption} 
\usepackage{algorithm}
\usepackage{algorithmic}

\usepackage{booktabs}
\usepackage{multirow}
\usepackage{enumitem}
\usepackage{newfloat}
\usepackage{listings}
\DeclareCaptionStyle{ruled}{labelfont=normalfont,labelsep=colon,strut=off} 
\floatstyle{ruled}
\newfloat{listing}{tb}{lst}{}
\floatname{listing}{Listing}
\title{Unveiling the Tapestry of Automated Essay Scoring: A Comprehensive Investigation of Accuracy, Fairness, and Generalizability}
\author{
    Kaixun Yang\textsuperscript{\rm 1},
    Mladen Rakovi\'{c}\textsuperscript{\rm 1},
    Yuyang Li\textsuperscript{\rm 1},
    Quanlong Guan\textsuperscript{\rm 2}\footnotemark[1],
    Dragan Ga\v{s}evi\'{c}\textsuperscript{\rm 1},
    Guanliang Chen\textsuperscript{\rm 1}\thanks{Corresponding authors.}
}
\affiliations{
    \textsuperscript{\rm 1}Centre for Learning Analytics at Monash, Monash University, Australia\\
    \textsuperscript{\rm 2}Department of Computer Science, Jinan University, China \\
    \{Kaixun.Yang1, Mladen.Rakovic, Dragan.Gasevic, Guanliang.Chen\}@monash.edu, 
    gql@jnu.edu.cn
}

\usepackage{bibentry}

\begin{document}

\maketitle

\begin{abstract}

Automatic Essay Scoring (AES) is a well-established educational pursuit that employs machine learning to evaluate student-authored essays. While much effort has been made in this area, current research primarily focuses on either (i) boosting the predictive accuracy of an AES model for a specific prompt (i.e., developing prompt-specific models), which often heavily relies on the use of the labeled data from the same target prompt; or (ii) assessing the applicability of AES models developed on non-target prompts to the intended target prompt (i.e., developing the AES models in a cross-prompt setting). Given the inherent bias in machine learning and its potential impact on marginalized groups, it is imperative to investigate whether such bias exists in current AES methods and, if identified, how it intervenes with an AES model's accuracy and generalizability. Thus, our study aimed to uncover the intricate relationship between an AES model's accuracy, fairness, and generalizability, contributing practical insights for developing effective AES models in real-world education. To this end, we meticulously selected nine prominent AES methods and evaluated their performance using seven distinct metrics on an open-sourced dataset, which contains over 25,000 essays and various demographic information about students such as gender, English language learner status, and economic status. Through extensive evaluations, we demonstrated that: (1) prompt-specific models tend to outperform their cross-prompt counterparts in terms of predictive accuracy; (2) prompt-specific models frequently exhibit a greater bias towards students of different economic statuses compared to cross-prompt models; (3) in the pursuit of generalizability, traditional machine learning models (e.g., SVM) coupled with carefully engineered features hold greater potential for achieving both high accuracy and fairness than complex neural network models.

\end{abstract}

\section{Introduction}\label{sec:intro}

In education, writing is a prevalent pedagogical practice employed by teachers and instructors to enhance student learning \cite{defazio2010academic}. Yet, the timely evaluation of students' essays or responses represents a formidable challenge, consuming considerable time and cognitive effort for educators. Recognizing the need to alleviate this burden, Automatic Essay Scoring (AES) has emerged, which refers to the process of using machine learning techniques to evaluate and assign scores to student-authored essays or responses \cite{chodorow2004beyond}. By automating this assessment process, educators can better focus on refining their teaching strategies, ultimately enabling a more efficient and effective learning experience for students.

Given the significant potential of AES, substantial efforts have been directed towards this field  \cite{ 10.1145/290941.290965, 10.1017/S1351324903003206, Chen2013AutomatedES, MCNAMARA201535}. It is important to highlight that a common objective shared among existing AES investigations is the pursuit of optimal predictive accuracy, i.e., correctly assessing and assigning scores to essays as many as possible. For instance, an early study \cite{zesch2015task} enhanced the training of an AES model based on Support Vector Machine (SVM) through a comprehensive feature set encompassing key linguistic attributes crucial for essay quality assessment (e.g., word n-gram features, cohesion features, and syntax features). The advancements in deep neural networks have spurred endeavors to further elevate predictive accuracy \cite{taghipour-ng-2016-neural, DBLP:journals/corr/AlikaniotisYR16}. These range from crafting dedicated scoring models based on different neural network architectures (e.g., Convolutional Neural Network (CNN) and Long Short-Term Memory (LSTM)) to harnessing pre-trained large language models (e.g., BERT \cite{devlin2019bert}). Noteworthy is that the aforementioned accuracy-focused studies were frequently operated within the \textit{prompt-specific} context, i.e., the AES models were developed and evaluated using labeled data exclusive to the intended target prompt. Nevertheless, obtaining such labeled data may not always be feasible, given its potential scarcity or the significant expenses and time required for its preparation. This has led to a recent trend in AES that centers on augmenting model generalizability in a \textit{cross-prompt} setting, i.e., building AES models based on pre-existing data sourced from non-target prompt and subsequently assessing their applicability to the desired target prompt \cite{jin2018tdnn, ridley2020prompt, li2020sednn}. 

While significant progress has been made, current research falls short in offering comprehensive insights to real-world educators on effectively balancing various factors crucial for constructing effective AES models. For instance, though the generalizability of cross-prompt models is a desirable trait to have, it is not the only trait that educators consider when determining which model they should use in practice. If the predictive accuracy of a cross-prompt model significantly lags behind that of a prompt-specific model, educators might favor the prompt-specific model, even when they acknowledge the associated costs of preparing tailored training data. However, the comparison between prompt-specific and cross-prompt AES models regarding their predictive accuracy remains largely unexplored in the existing literature. Beyond accuracy and generalizability, the predictive fairness of AES models has garnered increasing attention from educators. Here, predictive fairness entails ensuring that the essay score predictions generated by an AES model are impartial and unbiased across diverse student groups characterized by varying sensitive attributes such as gender and age. Undoubtedly, any bias hidden behind machine learning models can lead to unfair and discriminatory outcomes towards students, and thus should be addressed. Despite its recognized significance, the fairness of AES methods has received limited investigation within existing studies.

Hence, this study aimed to systematically investigate the intricate relationship between an AES model's accuracy and fairness, and generalizability, shedding light on practical insights for real-world educators to develop effective AES models to better support their teaching practices. Formally, this study was guided by the following two \textbf{R}esearch \textbf{Q}uestions:

\begin{enumerate}[label=\bfseries RQ\arabic*,leftmargin = 30pt]
    \item What is the performance difference between the prompt-specific and cross-prompt AES methods in terms of predictive \underline{\textit{accuracy}}?
    \item What is the performance difference between the prompt-specific and cross-prompt AES methods in terms of predictive \underline{\textit{fairness}}?
\end{enumerate}

To answer the RQs, we chose a publicly available dataset consisting of over 25,000 argumentative essays with holistic essay scores from 15 distinct prompts. Notably, this dataset provides various demographic details pertaining to students, including gender, economic status, disability status, English language learner status, and race. This rich dataset facilitated our exploration of AES model biases through various demographic lenses. To ensure a comprehensive evaluation, we extensively reviewed the existing AES literature and selected nine prominent methods in the field, with five from the prompt-specific category and four from the cross-prompt category. Subsequently, we replicated all nine methods and evaluate their accuracy and fairness measured by seven different metrics. The details are provided in Section Methods. We have publicly released our code\footnote{\url{https://github.com/CarsonYang518/AAAI24-AES-AFG}}.

In summary, this study contributed to the AES literature with the following main findings and insights:
\begin{itemize}
    \item Prompt-specific models tend to outperform their cross-prompt counterparts, with the performance gap ranging from 18.06\% to 25.61\% depending on the evaluation metrics used;
    \item In the cross-prompt setting, simple models (e.g., those based on well-investigated machine learning models like SVM) often excel in adequately identifying the characteristics of quality essays compared to complex models based on deep neural networks;
    \item Students' economic status emerges as the major attribute which frequently suffers from the predictive bias of existing AES models;
    \item Prompt-specific models frequently exhibit more bias towards students of different economic statuses compared to cross-prompt models;
    \item In the pursuit of generalizability, traditional machine learning models with carefully handcrafted features can achieve both high accuracy and fairness.
\end{itemize}

\section{Related Work}\label{sec:background}
\subsection{Automatic Essay Scoring}
The AES studies that are most relevant to our work can be broadly categorized into two groups, namely \textit{prompt-specific} and \textit{cross-prompt}, as briefly summarized below. 

\smallskip
\noindent\textbf{Prompt-specific AES.}
The initial explorations within this category predominantly relied on traditional machine learning techniques such as Bayesian Linear Regression, $\nu$-SVM, and Random Forests (RF) \cite{rudner2002automated, cozma2018automated, Chen2013AutomatedES}. To equip an AES model with the ability to accurately evaluate the quality of an essay, these studies often placed significant emphasis on the manual crafting of meaningful textual features as input to train the model. For instance, \cite{zesch2015task} empowered the training of a SVM-based scoring model by engineering an extensive set of linguistic features, encompassing critical aspects such as essay length, syntax, and coherence. Inspired by the strides made in deep learning techniques to address diverse natural language processing challenges, several studies have been dedicated to applying these methodologies to tackle AES \cite{taghipour-ng-2016-neural,dong2016automatic,dong2017attention,10.5555/3504035.3504765}. In contrast to traditional machine learning models, deep learning models dispense with the need for hand-crafted features, proficiently extracting such features from raw textual data. The focus of these deep learning studies often centers on the use of diverse deep neural network architectures that are adept at capturing distinct textual attributes within essays to facilitate subsequent grading. For example, CNNs are harnessed to discern local textual dependencies, while LSTM networks are employed to capture sequential dependencies \cite{taghipour-ng-2016-neural}. Hierarchical network structures are used to capture both word-level and sentence-level dependencies \cite{dong2016automatic}, and attention mechanisms are deployed to pinpoint pivotal words or sentences crucial for determining essay quality \cite{dong2017attention}. Besides, the recent advancements in pre-trained large language models (e.g., BERT \cite{devlin2019bert}) have spurred researchers to leverage these cutting-edge tools for automating essay assessment \cite{rodriguez2019language,yang2020enhancing,mizumoto2023exploring}. 

\smallskip
\noindent\textbf{Cross-prompt AES.} The common assumption held by this strand of studies is that, though the labeled data pertinent to the target prompt is unavailable to train a prompt-specific model, the quality of an essay can somewhat be revealed by features that are important across all prompts (e.g., the number of grammatical errors contained in the essay). Therefore, these studies often endeavored to craft such features to empower the training of an AES model \cite{zesch2015task, ridley2020prompt}. For example, \cite{zesch2015task} engineered weakly prompt-dependent features from  13 categories including the number of grammar errors, type-token-ratio, and readability score to train a SVM-based scoring model, whose predictive accuracy was up to 0.6856 measured by the metric of Quadratic Weighted Kappa (QWK). Building on top of this idea, \cite{jin2018tdnn} further proposed that the model built using the weakly prompt-dependent features could be used to accurately assign scores to certain essays, i.e., those receiving extremely high or low scores, and these essays together with their predicted scores can be used to further train a prompt-specific model. Specifically, \cite{jin2018tdnn} first trained a RankSVM model \cite{joachims2002optimizing} powered by weakly prompt-dependent features based on the labeled data collected from non-target prompts. Then, the RankSVM model was employed to identify a set of essays that were of extremely high or low scores from the target prompt, which were further used as input to train a prompt-specific model based on two-layer LSTM neural networks. Similar studies were presented in \cite{liu2021mfdnn,li2020sednn}. 

Despite substantial endeavors aimed at bolstering the generalizability of AES models, a comprehensive evaluation and comparison of the predictive accuracy disparity between cross-prompt models and prompt-specific counterparts remain absent in existing studies. This unavoidably hinders educators from understanding the inherent trade-off between accuracy and generalizability they might encounter when devising real-world AES models. More importantly, none of the aforementioned studies have undertaken an evaluation of the fairness aspect of existing AES models.

\subsection{Fair Machine Learning in Education}

Given the important role played by machine learning in supporting teaching and learning, an increasing amount of attention has been given to the predictive bias of machine learning techniques used in education. According to a recent survey \cite{li2023moral}, there have been only 49 peer-reviewed empirical papers on this topic published after 2010. These papers mostly centered around the tasks such as predicting students' course performance or their likelihood of dropping out from a course. To our knowledge, there are only two papers that diagnosed the predictive bias displayed by AES models, even though the importance of this task has been pointed out as early as in 2012 \cite{williamson2012framework}. Specifically, \cite{litman2021fairness} evaluated the fairness of three prompt-specific models, i.e., one based on the RF model with handcrafted features, one based on CNN-LSTM-Attention with textual features, and one based on CNN-LSTM-Attention with hybrid features. Nonetheless, this study is limited in that it did not include any cross-prompt models and the findings were derived based on a private dataset consisting of data from only one prompt, which inherently hinders their reproducibility and generalizability in similar scenarios. In contrast, we delivered a more comprehensive evaluation by including nine prominent methods that encompass both the prompt-specific and cross-prompt settings, and the evaluation was based on a larger-scale public dataset collected from 15 distinct prompts. Additionally, \cite{doewes2022individual} measured individual fairness in AES while our work focused on group fairness.
 
\section{Methods} \label{methods}
\subsection{Datasets}
A major obstacle hindering the exploration of fair AES is the absence of demographic information within widely used datasets for AES research. To our knowledge, only two public datasets contain students' demographic information: the ELLIPSE Corpus and the PERSUADE 2.0 corpus \footnote{\url{https://github.com/scrosseye/persuade_corpus_2.0}}. For our evaluation, we chose the PERSUADE 2.0 corpus due to its larger dataset size, approximately five times that of the ELLIPSE Corpus. This will adequately fulfill the requirements of training data quantity for AES models based deep learning techniques, as described in Section Models.

The PERSUADE 2.0 corpus originally consists of over 25,000 argumentative essays written by students from 6th to 12th grade in the US for 15 different prompts \cite{crossley2022persuasive}. The holistic essay scores, which serve as the ground truth for this study's predictions, were assigned by human raters who underwent training on a scoring rubric employed in the standardized Scholastic Aptitude Test (SAT) in the US. These holistic scores span from 1 to 6, denoting low to high quality, with increments of 1. Importantly, the dataset encompasses five demographic attributes of the students, including \textit{gender} (male vs. female), \textit{race/ethnicity} (e.g., White, Asian, Black), \textit{economic status} (economically disadvantaged vs. non-economically disadvantaged), \textit{English language learner status} (native English speakers vs. non-native English speakers), and \textit{disability status} (students with disabilities and those without). All these demographic attributes were considered to address RQ2. The details of the dataset are described in \cite{crossley_scott_2023_8221504}. 

\subsection{Models} \label{sec:models}
Following prior research \cite{10.5555/3504035.3504765, ridley2020prompt, jin2018tdnn, cozma2018automated, yang2020enhancing}, we treated the prediction of an essay's score as a regression problem. To ensure a comprehensive evaluation, we conducted an extensive review on the existing AES literature, after which we chose five representative \textbf{\textit{prompt-specific}} methods, as described below:

\begin{itemize}

\item \textbf{\texttt{SVM (Full)}} \cite{zesch2015task}, which aims to adequately empower the training of a SVM-based scoring model by carefully engineering a comprehensive set of features from raw essay text. The authors distinguished two types of features, namely (i) \textit{strongly prompt-dependent} ones, i.e., those highly associated with a specific prompt such as word n-grams and essay length; and (ii) \textit{weakly prompt-dependent} ones, i.e., those matter to the essay assessment of all prompts such as grammatical errors and readability.

\item \textbf{\texttt{SKIPFLOW-LSTM}} \cite{10.5555/3504035.3504765}, which is a pioneering attempt to incorporate features related to the coherence of an essay (i.e., the semantic similarity between different sentences) to train an AES based on an end-to-end neural network architecture. This architecture encompasses a neural tensor layer to capture the relationship between two LSTM outputs, with the goal of automatically extracting coherence features for essay scoring.

\item \textbf{\texttt{CNN-LSTM-ATT}} \cite{dong2017attention}, which is the first study to employ a neural hierarchical sentence-document architecture for AES. Specifically, this study used CNN to capture the word relations in a sentence and then LSTM to capture the sentence relations in an essay. Besides, the attention mechanism was applied to identify crucial words and sentences for assessing essay quality.

\item \textbf{\texttt{R$^2$BERT}} \cite{yang2020enhancing}, which is a pioneering attempt to combine the methodologies of regression and ranking in AES. Specifically, a hybrid loss with the dynamic weights of mean square error loss (i.e., regression loss) and batch-wise ListNet loss (i.e., ranking loss) is applied to fine-tune the BERT for AES.

\item \textbf{\texttt{BERT (3 Layers)}} \cite{rodriguez2019language}, which aims to mitigate overfitting by only using the initial three layers of BERT to produce essay representations for subsequent scoring. This configuration setting was demonstrated to yield the optimal performance after extensive experimentation of various alternative configurations and training techniques for BERT.

\end{itemize}

Similarly, we chose four representative \textbf{\textit{cross-prompt}} methods, as described below:

\begin{itemize}
\item \textbf{\texttt{SVM (Reduced)}} \cite{zesch2015task}, which is similar to \texttt{SVM (Full)} described above, but only using the weakly prompt-dependent features for model training.

\item \textbf{\texttt{RankSVM}} \cite{chen2014automated}, which is a representative ranking-based method for AES. A RankSVM is first trained using pair-wise essays ordered by ground-truth scores. Then, the constructed RankSVM is used to generate intermediate scores for ranking the essays, and such intermediate scores are subsequently mapped to a pre-defined scoring scale to generate the essay scores.

\item \textbf{\texttt{PAES}} \cite{ridley2020prompt}, which is similar to \texttt{CNN-LSTM-ATT} mentioned above. However, Part-of-Speech (POS) embeddings are used here rather than word embeddings, because POS embeddings are assumed to be more effective in generating a generalized representation of an essay. Besides, this method incorporates certain weakly task-dependent features to train the AES model.

\item \textbf{\texttt{TDNN}} \cite{jin2018tdnn}, which introduces a pioneering two-step approach for cross-prompt essay scoring. Firstly, a RankSVM model is constructed as described above. Secondly, the RankSVM model is used to assign scores to essays in the desired target prompt, among which the essays receiving extremely high or low scores are further used to train a LSTM neural network for prompt-specific essay scoring.

\end{itemize}

\begin{table*}[hbt!]
\begin{center}
\resizebox{1\textwidth}{!}{
\begin{tabular}{@{}l|l|lll|lll|lll|lll@{}}
\toprule
\multirow{2}{*}{\textbf{\begin{tabular}[c]{@{}l@{}}Types\end{tabular}}} &
  \textbf{Metrics} &
  \textbf{$\uparrow$ QWK} &
  \textbf{$\downarrow$ MAE} &
  \textbf{$\uparrow$ PCC} &
  \textbf{$\uparrow$ QWK} &
  \textbf{$\downarrow$ MAE} &
  \textbf{$\uparrow$ PCC} &
  \textbf{$\uparrow$ QWK} &
  \textbf{$\downarrow$ MAE} &
  \textbf{$\uparrow$ PCC} &
  \textbf{$\uparrow$ QWK} &
  \textbf{$\downarrow$ MAE} &
  \textbf{$\uparrow$ PCC} \\ \cmidrule(l){2-14}
 &
  \textbf{Methods} &
  \multicolumn{3}{c|}{\textbf{Prompt 1}} &
  \multicolumn{3}{c|}{\textbf{Prompt 2}} &
  \multicolumn{3}{c|}{\textbf{Prompt 3}} &
  \multicolumn{3}{c}{\textbf{Prompt 4}} \\ \midrule
\multirow{5}{*}{PS} &
  SVM (Full) &
  0.823 &
  0.456 &
  0.845 &
  0.770 &
  0.495 &
  0.819 &
  0.772 &
  0.528 &
  0.816 &
  0.717 &
  0.473 &
  0.758 \\
 &
  SKIPFLOW-LSTM &
  0.781 &
  0.527 &
  0.798 &
  0.838 &
  0.465 &
  0.849 &
  0.800 &
  0.532 &
  0.813 &
  0.713 &
  0.489 &
  0.747 \\
 &
  CNN-LSTM-ATT &
  0.838 &
  0.473 &
  0.846 &
  0.855 &
  0.451 &
  0.864 &
  0.824 &
  0.515 &
  0.832 &
  0.777 &
  \textbf{0.444} &
  \textbf{0.796} \\
 &
  R$^2$BERT &
  0.822 &
  0.530 &
  0.826 &
  0.849 &
  \textbf{0.422} &
  0.857 &
  0.831 &
  \textbf{0.469} &
  0.837 &
  0.740 &
  0.464 &
  0.751 \\
 &
  BERT (3 Layers) &
  \textbf{0.858} &
  \textbf{0.452} &
  \textbf{0.861} &
  \textbf{0.866} &
  0.435 &
  \textbf{0.869} &
  \textbf{0.844} &
  0.501 &
  \textbf{0.850} &
  \textbf{0.787} &
  0.451 &
  0.792 \\ \midrule
\multirow{4}{*}{CP} &
  SVM (Reduced) &
  0.835 &
  0.462 &
  0.851 &
  0.774 &
  0.520 &
  0.806 &
  0.761 &
  0.566 &
  0.785 &
  0.759 &
  0.475 &
  0.773 \\
 &
  RankSVM &
  0.747 &
  0.573 &
  0.798 &
  0.587 &
  0.710 &
  0.702 &
  0.499 &
  0.820 &
  0.678 &
  0.584 &
  0.602 &
  0.667 \\
 &
  PAES &
  0.809 &
  0.567 &
  0.833 &
  0.760 &
  0.566 &
  0.783 &
  0.730 &
  0.669 &
  0.754 &
  0.666 &
  0.719 &
  0.719 \\
 &
  TDNN &
  0.732 &
  0.692 &
  0.789 &
  0.609 &
  0.690 &
  0.646 &
  0.560 &
  0.721 &
  0.642 &
  0.536 &
  0.602 &
  0.599 \\ \midrule \midrule
\textbf{} &
   &
  \multicolumn{3}{c|}{\textbf{Prompt 5}} &
  \multicolumn{3}{c|}{\textbf{Prompt 6}} &
  \multicolumn{3}{c|}{\textbf{Prompt 7}} &
  \multicolumn{3}{c}{\textbf{Prompt 8}} \\ \midrule
\multirow{5}{*}{PS} &
  SVM (Full) &
  0.753 &
  0.475 &
  0.790 &
  0.783 &
  0.449 &
  0.813 &
  0.732 &
  0.425 &
  0.774 &
  0.704 &
  0.454 &
  0.747 \\
 &
  SKIPFLOW-LSTM &
  0.674 &
  0.552 &
  0.708 &
  0.773 &
  0.487 &
  0.793 &
  0.730 &
  0.451 &
  0.752 &
  0.728 &
  0.451 &
  0.745 \\
 &
  CNN-LSTM-ATT &
  0.775 &
  0.477 &
  \textbf{0.793} &
  0.813 &
  0.449 &
  0.826 &
  0.776 &
  0.414 &
  0.795 &
  0.763 &
  0.434 &
  0.779 \\
 &
  R$^2$BERT &
  0.708 &
  0.556 &
  0.727 &
  0.802 &
  0.473 &
  0.815 &
  0.783 &
  0.389 &
  0.788 &
  0.768 &
  0.433 &
  0.772 \\
 &
  BERT (3 Layers) &
  \textbf{0.783} &
  0.486 &
  0.792 &
  \textbf{0.840} &
  \textbf{0.428} &
  \textbf{0.845} &
  \textbf{0.806} &
  \textbf{0.393} &
  \textbf{0.811} &
  \textbf{0.788} &
  \textbf{0.417} &
  \textbf{0.793} \\ \midrule
\multirow{4}{*}{CP} &
  SVM (Reduced) &
  0.764 &
  \textbf{0.474} &
  0.789 &
  0.822 &
  0.437 &
  0.833 &
  0.747 &
  0.438 &
  0.790 &
  0.740 &
  0.436 &
  0.771 \\
 &
  RankSVM &
  0.691 &
  0.551 &
  0.718 &
  0.723 &
  0.517 &
  0.743 &
  0.521 &
  0.537 &
  0.546 &
  0.522 &
  0.581 &
  0.563 \\
 &
  PAES &
  0.690 &
  0.661 &
  0.751 &
  0.749 &
  0.606 &
  0.809 &
  0.711 &
  0.548 &
  0.747 &
  0.662 &
  0.620 &
  0.742 \\
 &
  TDNN &
  0.620 &
  0.608 &
  0.641 &
  0.727 &
  0.581 &
  0.733 &
  0.090 &
  0.849 &
  0.157 &
  0.225 &
  0.743 &
  0.317 \\ \midrule \midrule
\textbf{} &
   &
  \multicolumn{3}{c|}{\textbf{Prompt 9}} &
  \multicolumn{3}{c|}{\textbf{Prompt 10}} &
  \multicolumn{3}{c|}{\textbf{Prompt 11}} &
  \multicolumn{3}{c}{\textbf{Prompt 12}} \\ \midrule
\multirow{5}{*}{PS} &
  SVM (Full) &
  0.719 &
  0.486 &
  0.778 &
  0.756 &
  0.431 &
  0.791 &
  0.713 &
  0.457 &
  0.755 &
  0.557 &
  0.464 &
  0.686 \\
 &
  SKIPFLOW-LSTM &
  0.765 &
  0.482 &
  0.785 &
  0.697 &
  0.487 &
  0.742 &
  0.702 &
  0.475 &
  0.732 &
  0.672 &
  0.567 &
  0.697 \\
 &
  CNN-LSTM-ATT &
  0.793 &
  0.461 &
  0.816 &
  0.784 &
  0.427 &
  0.804 &
  0.761 &
  0.435 &
  0.783 &
  0.708 &
  0.565 &
  0.731 \\
 &
  R$^2$BERT &
  0.778 &
  0.465 &
  0.793 &
  0.735 &
  0.470 &
  0.742 &
  0.761 &
  0.434 &
  0.772 &
  0.750 &
  \textbf{0.369} &
  0.755 \\
 &
  BERT (3 Layers) &
  \textbf{0.813} &
  \textbf{0.453} &
  \textbf{0.819} &
  \textbf{0.802} &
  \textbf{0.414} &
  \textbf{0.811} &
  \textbf{0.794} &
  \textbf{0.413} &
  \textbf{0.800} &
  \textbf{0.750} &
  0.437 &
  \textbf{0.756} \\ \midrule
\multirow{4}{*}{CP} &
  SVM (Reduced) &
  0.621 &
  0.621 &
  0.715 &
  0.802 &
  0.424 &
  0.806 &
  0.749 &
  0.432 &
  0.784 &
  0.557 &
  0.494 &
  0.599 \\
 &
  RankSVM &
  0.562 &
  0.623 &
  0.607 &
  0.725 &
  0.497 &
  0.729 &
  0.470 &
  0.662 &
  0.598 &
  0.319 &
  0.677 &
  0.362 \\
 &
  PAES &
  0.524 &
  0.861 &
  0.622 &
  0.637 &
  0.794 &
  0.735 &
  0.679 &
  0.589 &
  0.738 &
  0.540 &
  0.511 &
  0.573 \\
 &
  TDNN &
  0.320 &
  0.749 &
  0.472 &
  0.671 &
  0.656 &
  0.691 &
  0.108 &
  0.722 &
  0.341 &
  0.199 &
  0.632 &
  0.303 \\ \bottomrule
\end{tabular}
}
\end{center}
 \vspace{-3mm}
\caption{The predictive accuracy of the selected AES methods in each prompt. PS represents Prompt-Specific. CP represents Cross-Prompt. Bold values represent the best performance in a metric. The signs $\uparrow$ and $\downarrow$ indicate whether a higher ($\uparrow$) or lower ($\downarrow$) value is more preferred in a metric.}
\label{table1} \vspace{-3mm}
\end{table*}

\begin{table*}[hbt!]
\begin{center}
\resizebox{1\textwidth}{!}
{
\begin{tabular}{@{}l|l|cccc|cccc|cccc|cccc@{}}
\toprule
\multirow{2}{*}{\textbf{Types}} & \textbf{Metrics} & \multicolumn{1}{l}{\textbf{OSA}} & \multicolumn{1}{l}{\textbf{OSD}} & \multicolumn{1}{l}{\textbf{CSD}} & \multicolumn{1}{l|}{\textbf{MAED}} & \multicolumn{1}{l}{\textbf{OSA}} & \multicolumn{1}{l}{\textbf{OSD}} & \multicolumn{1}{l}{\textbf{CSD}} & \multicolumn{1}{l|}{\textbf{MAED}} & \multicolumn{1}{l}{\textbf{OSA}} & \multicolumn{1}{l}{\textbf{OSD}} & \multicolumn{1}{l}{\textbf{CSD}} & \multicolumn{1}{l|}{\textbf{MAED}} & \multicolumn{1}{l}{\textbf{OSA}} & \multicolumn{1}{l}{\textbf{OSD}} & \multicolumn{1}{l}{\textbf{CSD}} & \multicolumn{1}{l}{\textbf{MAED}} \\ \cmidrule(l){2-18} 
 & \textbf{Methods} & \multicolumn{4}{c|}{\textbf{Prompt 1}} & \multicolumn{4}{c|}{\textbf{Prompt 2}} & \multicolumn{4}{c|}{\textbf{Prompt 3}} & \multicolumn{4}{c}{\textbf{Prompt 4}} \\ \midrule
\multirow{5}{*}{PS} & SVM (Full) & ns & ns & ns & 0.052 & ns & 0.022 & ns & -0.087 & ns & 0.045 & ns & -0.011 & ns & ns & ns & -0.008 \\
 & SKIPFLOW-LSTM & 0.111 & 0.112 & 0.014 & -0.512 & 0.078 & 0.079 & 0.005 & -0.459 & 0.051 & 0.052 & ns & -0.386 & 0.043 & 0.043 & ns & -0.243 \\
 & CNN-LSTM-ATT & ns & ns & 0.020 & 0.045 & ns & ns & 0.016 & -0.043 & ns & ns & ns & 0.033 & ns & ns & ns & 0.012 \\
 & R$^2$BERT & ns & ns & 0.022 & 0.010 & ns & ns & 0.023 & -0.084 & ns & ns & ns & 0.001 & ns & ns & ns & -0.031 \\
 & BERT (3 Layers) & ns & ns & 0.015 & 0.037 & ns & ns & ns & -0.029 & ns & ns & ns & 0.015 & ns & ns & ns & -0.005 \\ \midrule
\multirow{4}{*}{CP} & SVM (Reduced) & ns & ns & ns & 0.015 & ns & ns & ns & 0.022 & ns & 0.031 & ns & 0.035 & ns & ns & ns & 0.001 \\
 & Rank-SVM & ns & ns & ns & 0.016 & ns & 0.025 & ns & 0.107 & 0.015 & 0.05 & ns & 0.180 & ns & ns & ns & 0.078 \\
 & PAES & ns & ns & ns & 0.087 & ns & ns & ns & 0.042 & ns & ns & ns & 0.048 & ns & ns & 0.028 & -0.012 \\
 & TDNN & ns & ns & ns & 0.099 & ns & 0.022 & ns & 0.057 & ns & 0.054 & ns & 0.036 & ns & ns & 0.005 & 0.033 \\ \midrule \midrule
\textbf{} &  & \multicolumn{4}{c|}{\textbf{Prompt 5}} & \multicolumn{4}{c|}{\textbf{Prompt 6}} & \multicolumn{4}{c|}{\textbf{Prompt 7}} & \multicolumn{4}{c}{\textbf{Prompt 8}} \\ \midrule
\multirow{5}{*}{PS} & SVM (Full) & ns & ns & ns & 0.031 & ns & ns & ns & 0.040 & ns & ns & ns & -0.037 & ns & ns & ns & -0.014 \\
 & SKIPFLOW-LSTM & 0.051 & 0.051 & ns & -0.293 & 0.104 & 0.111 & ns & -0.494 & ns & ns & ns & -0.239 & 0.038 & 0.04 & ns & -0.245 \\
 & CNN-LSTM-ATT & ns & ns & ns & 0.022 & ns & ns & 0.012 & 0.008 & ns & ns & ns & -0.008 & ns & ns & ns & -0.003 \\
 & R$^2$BERT & ns & ns & ns & -0.001 & ns & ns & 0.022 & -0.035 & ns & ns & ns & -0.014 & ns & ns & ns & -0.002 \\
 & BERT (3 Layers) & ns & ns & ns & 0.038 & ns & ns & 0.026 & -0.023 & ns & ns & ns & -0.008 & ns & ns & 0.015 & -0.018 \\ \midrule
\multirow{4}{*}{CP} & SVM (Reduced) & ns & ns & ns & 0.026 & ns & ns & ns & 0.016 & ns & ns & ns & -0.038 & ns & ns & ns & -0.006 \\
 & Rank-SVM & ns & ns & ns & 0.001 & ns & ns & ns & 0.090 & ns & ns & ns & 0.017 & ns & 0.015 & ns & 0.066 \\
 & PAES & ns & ns & ns & 0.072 & ns & ns & ns & 0.084 & ns & ns & 0.022 & -0.005 & ns & ns & 0.009 & -0.036 \\
 & TDNN & ns & ns & ns & 0.041 & ns & ns & ns & 0.012 & ns & 0.048 & ns & -0.221 & ns & 0.03 & ns & -0.149 \\ \midrule \midrule
\textbf{} &  & \multicolumn{4}{c|}{\textbf{Prompt 9}} & \multicolumn{4}{c|}{\textbf{Prompt 10}} & \multicolumn{4}{c|}{\textbf{Prompt 11}} & \multicolumn{4}{c}{\textbf{Prompt 12}} \\ \midrule
\multirow{5}{*}{PS} & SVM (Full) & ns & ns & ns & -0.023 & ns & ns & ns & 0.006 & ns & ns & ns & -0.002 & ns & ns & ns & -0.022 \\
 & SKIPFLOW-LSTM & 0.036 & 0.038 & ns & -0.259 & 0.05 & 0.05 & ns & -0.265 & 0.016 & 0.019 & ns & -0.163 & ns & ns & ns & -0.189 \\
 & CNN-LSTM-ATT & ns & ns & ns & -0.001 & ns & ns & 0.007 & -0.003 & ns & ns & ns & 0.014 & ns & ns & ns & -0.027 \\
 & R$^2$BERT & ns & ns & ns & -0.030 & ns & ns & ns & -0.058 & ns & ns & ns & -0.048 & ns & ns & ns & -0.031 \\
 & BERT (3 Layers) & ns & ns & ns & -0.043 & ns & ns & ns & 0.015 & ns & ns & ns & -0.009 & ns & ns & ns & -0.029 \\ \midrule
\multirow{4}{*}{CP} & SVM (Reduced) & ns & ns & ns & -0.059 & ns & ns & ns & 0.006 & ns & ns & ns & 0.009 & ns & ns & ns & -0.016 \\
 & Rank-SVM & ns & ns & ns & -0.002 & ns & ns & ns & 0.003 & 0.02 & 0.023 & ns & 0.153 & ns & ns & ns & 0.042 \\
 & PAES & ns & ns & ns & -0.081 & ns & ns & ns & 0.054 & ns & ns & ns & -0.014 & ns & ns & ns & 0.003 \\
 & TDNN & ns & ns & ns & -0.133 & ns & ns & ns & 0.015 & ns & 0.052 & ns & 0.102 & ns & ns & ns & 0.001 \\ \bottomrule
\end{tabular}
}
\end{center}
 \vspace{-3mm}
\caption{The predictive fairness of the selected AES methods for Economic Status. S represents Prompt-Specific. CP represents Cross-Prompt. The `ns' label indicates non-significant results (p $<$ 0.05). Lower values indicate a higher level of fairness.} 
\label{table2} 
\vspace{-3mm}
\end{table*}

\subsection{Experimental Setup}

\smallskip
\noindent\textbf{Data Preprocessing.}
The essays without corresponding student demographic information were removed, which resulted in a total of 20,626 essays spanning 12 distinct prompts for our evaluation. Notably, four out of the five demographic attributes are in a binary form (e.g., male vs. female), except for \textit{race/ethnicity}, which contains six values including White, Hispanic, Black, Asian, American Indian, and Other. As guided by \cite{hutt2019evaluating, bayer2021learning}, White students are regarded as the privileged group, we converted this attribute to binary values of White vs. Non-White. In line with existing studies in the field \cite{litman2021fairness}, we treated students who are either male, White, economically advantaged, native English speakers, or without disabilities as the privileged group and the others as the non-privileged group to measure the fairness of AES models.

\smallskip
\noindent\textbf{Feature Engineering.}
The handcrafted features of the models \texttt{SVM (Full/Reduced)}, \texttt{RankSVM}, \texttt{PAES}, and \texttt{TDNN} (as specified in Section Models were derived using NLTK \cite{loper2002nltk}, Stanza \cite{qi2020stanza}, and spaCy \cite{honnibal2017spacy}. In line with previous studies \cite{jin2018tdnn, ridley2020prompt}, we standardized all handcrafted features, adjusting their means to 0 and standard deviations to 1.0. The details about handcrafted features can be found in the Appendix.

\smallskip
\noindent\textbf{Model Construction.}
We employed both TensorFlow \cite{paszke2019pytorch} and PyTorch \cite{tensorflow2015-whitepaper} frameworks for implementing the deep learning models, namely \texttt{SKIPFLOW-LSTM}, \texttt{CNN-LSTM-ATT}, \texttt{R$^2$BERT}, \texttt{BERT (3 Layers)}, \texttt{PAES}, and \texttt{TDNN} (as detailed in Section Models). For traditional machine learning models (e.g., SVM), we employed Scikit-learn \cite{scikit-learn}. For RankSVM, we employed the SVMs library \footnote{\url{https://www.cs.cornell.edu/people/tj/svm_light/svm_rank.html}}. All the model hyperparameters were set following the guidelines specified in the original papers and can be found in the Appendix. All the model training and evaluation were performed on Google Colab Pro with 16 GB of RAM and an NVIDIA Tesla T4 GPU.

\smallskip
\noindent\textbf{Evaluation Procedure.}
In previous studies \cite{mathias2018asap++, cozma2018automated, dong2017attention, MCNAMARA201535}, the evaluation of prompt-specific methods often involved 5-fold cross-validation. As for the evaluation of cross-prompt methods, a prompt-wise cross-validation approach is commonly employed \cite{ridley2020prompt, jin2018tdnn, liu2021mfdnn}, where essays corresponding to a target prompt are held out for testing, while the remaining essays of other prompt are utilized as training data. We adopt the same evaluation procedure as previous studies. By doing this, all the essays contained in a target prompt were scored by prompt-specific and cross-prompt methods, which enabled us to directly compare their performance.

\subsection{Evaluation Metric}
To measure \textit{accuracy}, we adopt three commonly used metrics in existing AES literature \cite{lagakis2021automated}: Quadratic Weighted Kappa (QWK), Mean Absolute Error (MAE), and Pearson Correlation Coefficient (PCC). Although QWK is designed for categorical variables, we adapted it for our regression task by utilizing a modified version suited for continuous values \cite{haberman2019measures}.

To measure \textit{fairness}, we aligned with previous studies \cite{loukina2019many, litman2021fairness} and adopt three metrics to measure to what extent the predictive errors of an AES model towards different student groups can be attributed to their demographic traits:

\begin{itemize}
    \item Overall Score Accuracy (OSA), which measures the parity of an AES model in terms of the variance between its predicted scores and the ground-truth scores that can be explained by students' demographic attributes. Specifically, OSA represents the scores given by an AES model and the human rater with $S$ and $H$, respectively. Then, a linear regression is constructed with $(S-H)^2$ as the dependent variable and demographic attributes as the independent variable. OSA is calculated as the $R^2$ of this regression model.
    \item Overall Score Difference (OSD), which is similar to OSA, but with $S-H$ (instead of $(S-H)^2$) to construct the regression model. This is designed to capture any ``overestimation'' or ``underestimation'' displayed by an AES model towards any group of students (e.g., whether the AES model tends to assign higher scores to essays written by male students while their female counterparts often receive lower scores).
    \item Conditional Score Difference (CSD), which is similar to OSD, takes a step further by accounting for students' language proficiency, approximated by their ground-truth essay scores. This is achieved by constructing two regression models with $S-H$ as the dependent variable, first with $H$ as the independent variable, and then with both $H$ and demographic attributes. CSD is calculated as the difference between $R^2$ of these two regression models.    
\end{itemize}

The larger the OSA/OSD/CSD, the more bias an AES model has. We employed ANOVA to assess whether the results of CSD were statistically significant. In addition to using OSA, OSD, and CSD to explain the scoring error variance across different demographic groups, we measured fairness from the scale perspective by adopting Mean Absolute Error Difference (MAED) \cite{sun2022processing}, which calculates the difference between the MAE of the privileged and unprivileged groups. Positive MAED values indicate that the AES model holds bias towards the privileged group while negative values indicate bias towards the non-privileged group. That is, the closer a MAED is to 0, the more fair an AES model is. All the evaluation metrics were calculated using RSMTool \cite{MadnaniLoukina2016}.

\section{Results} \label{sec:res}
\subsection{Results on RQ1} \label{subsec:accuracy_results}

The predictive accuracy of the nine selected AES methods in each prompt is detailed in Table \ref{table1}, which is further averaged and presented in Table \ref{table3}. Based on these tables, two interesting observations can be made.

Firstly, prompt-specific models generally outperform cross-prompt models. As shown in Table \ref{table3}, on average, QWK exhibits a 25.61\% increase, the MAE shows a reduction of 23.43\%, and the PCC demonstrates an enhancement of 18.06\%. When comparing the best-performing prompt-specific model \texttt{BERT (3 Layers)})  and its best-performing cross-prompt counterpart (i.e., \texttt{SVM (Reduced)}), the performance gap is 9.00\% in QWK, 8.71\% in MAE, and 5.42\% in PCC. On the other hand, in line with previous research \cite{zesch2015task, cozma2018automated}, we observed that prompt-specific models tended to display greater robustness compared to cross-prompt ones, as evidenced by the variances shown in Table \ref{table3}. This is due to the more challenging nature of the cross-prompt essay scoring as it can not leverage prompt-specific features (e.g., n-grams) that directly contribute to the accurate evaluation of an essay. 

Secondly, when scrutinizing the prompt-specific models, we observe that models based on deep neural networks are consistently superior to those based on traditional machine learning techniques. For instance, the best performing model \texttt{BERT (3 Layers)} achieved an average performance of up to 0.811 (QWK), 0.440 (MAE), and 0.817 (PCC). Notably, this model also achieved the highest level of robustness as indicated by the lowest variances (as low as 0.001) among all the prompt-specific models. However, when scrutinizing the cross-prompt models, we have the contrary finding, i.e., the traditional machine learning method \texttt{SVM (Reduced)} exhibits the highest performance compared to all the other deep learning methods (namely PAES and TDNN). This implies that, in the cross-prompt setting, simple models can effectively discern significant patterns of quality essays by using weakly prompt-dependent features, while complex models based on deep neural networks (e.g., TDNN, which is an advanced version of RankSVM) have the tendency to overfit non-target-prompt essays, thereby diminishing their ability to generalize effectively. 

\subsection{Results on RQ2}
\begin{table*}[!htb]
 \begin{center}
 \resizebox{0.8\textwidth}{!}{
\begin{tabular}{@{}l|l|cc|cc|cc|cccc@{}}
\toprule
\textbf{Types} & \textbf{Methods} & \multicolumn{1}{l}{$\uparrow$\textbf{QWK}} & \multicolumn{1}{l|}{\textbf{$\sigma^2$}} & \multicolumn{1}{l}{$\downarrow$\textbf{MAE}} & \multicolumn{1}{l|}{\textbf{$\sigma^2$}} & \multicolumn{1}{l}{$\uparrow$\textbf{PCC}} & \multicolumn{1}{l|}{\textbf{$\sigma^2$}} & \multicolumn{1}{l}{\textbf{OSA}} & \multicolumn{1}{l}{\textbf{OSD}} & \multicolumn{1}{l}{\textbf{CSD}} & \multicolumn{1}{l}{\textbf{MAED}} \\ \midrule
\multirow{5}{*}{PS} & SVM(Full) & 0.733 & 0.004 & 0.466 & 0.001 & 0.781 & 0.002 & 0 & 2 & 0 & -0.006 \\ \cmidrule(l){2-12} 
 & SKIPFLOW-LSTM & 0.739 & 0.003 & 0.497 & 0.001 & 0.763 & 0.002 & 10 & 10 & 2 & -0.312 \\ \cmidrule(l){2-12} 
 & CNN-LSTM-ATT & 0.789 & 0.002 & 0.462 & 0.002 & 0.805 & 0.001 & 0 & 0 & 4 & 0.004 \\ \cmidrule(l){2-12} 
 & R$^2$BERT & 0.777 & 0.002 & 0.456 & 0.003 & 0.786 & 0.002 & 0 & 0 & 3 & -0.027 \\ \cmidrule(l){2-12} 
 & BERT(3 Layers) & \textbf{0.811} & 0.001 & \textbf{0.440} & 0.001 & \textbf{0.817} & 0.001 & 0 & 0 & 3 & -0.005 \\ \midrule
\multirow{4}{*}{CP} & SVM(Reduced) & 0.744 & 0.006 & 0.482 & 0.004 & 0.775 & 0.004 & 0 & 1 & 0 & 0.001 \\ \cmidrule(l){2-12} 
 & Rank-SVM & 0.579 & 0.016 & 0.613 & 0.009 & 0.643 & 0.014 & 2 & 4 & 0 & 0.063 \\ \cmidrule(l){2-12} 
 & PAES & 0.680 & 0.007 & 0.643 & 0.011 & 0.734 & 0.005 & 0 & 0 & 3 & 0.020 \\ \cmidrule(l){2-12} 
 & TDNN & 0.450 & 0.060 & 0.687 & 0.006 & 0.528 & 0.041 & 0 & 4 & 2 & -0.009 \\ \bottomrule
\end{tabular}
}
 \end{center}
  \vspace{-3mm}
 \caption{The average accuracy performance and overall fairness performance for Economic Status of the selected AES methods across all prompts. $\sigma^2$ represents variance. PS represents Prompt-Specific. CP represents Cross-Prompt. Bold values represent the best performance in a metric. The signs $\uparrow$ and $\downarrow$ indicate whether a higher ($\uparrow$) or lower ($\downarrow$) value is more preferred in a metric. Cells in OSA, OSD, and CSD denote the number of prompts in which an AES method was diagnosed to have predictive bias, e.g., the number of cells with values other than `ns' in Table \ref{table2}. Cells in  MAED represent the average MAED of all prompts.} \label{table3} 
 \vspace{-3mm}
\end{table*}

The predictive fairness was evaluated by using all the five available demographic attributes, among which we observed that an AES model's bias is frequently associated with a student's \textit{economic status}. The predictive bias of AES methods in different prompts is given in Table \ref{table2}, which are further summarized and presented in Table \ref{table3}. This aligns with the findings presented in previous studies \cite{abdu2015parents}, i.e., there exists a relationship between students' academic achievements and their parents' socio-economic status. On the other hand, \textit{gender} is the attribute in which AES models display relatively fewer biases in our case. Due to the limited space, the results of the other four demographic attributes are given in the Appendix.

When delving into the results of economic status presented in Table \ref{table2} and Table \ref{table3}, we observe that prompt-specific models generally displayed more bias compared to their cross-prompt counterparts. Specifically, when centering on the metrics of OSA, OSD, and CSD, the average number of prompts that the prompt-specific models were diagnosed to have bias is greater than that of the cross-prompt models, namely 2.0 vs. 0.5 in OSA, 2.4 vs. 2.25 in OSD, and 2.4 vs. 1.25 in CSD. On the other hand, when calculating the average of the absolute MAED values, the performance of prompt-specific models is 2.63\% higher than that of the cross-prompt models. It should be noted that cross-prompt models tended to favor the non-privileged group (i.e., three out of four models displayed positive MAED values) while prompt-specific models were more likely to favor the privileged group (i.e., four out of the five models displayed negative MAED values).

When scrutinizing the fairness displayed by individual models in the prompt-specific setting, we observe that \texttt{SVM (Full)} is superior to the other methods, with only two prompts detected with bias measured in OSD and a minimal MAED value of -0.006. This model is followed by \texttt{BERT (3 Layers)}, which was diagnosed to be biased in three prompts and with a MAED value of -0.005. Recall the RQ1 results presented in Table \ref{table1} and Table \ref{table3}, \texttt{BERT (3 Layers)} demonstrated the highest predictive accuracy and robustness. This further strengthens the superiority of AES models based on meticulously fine-tuned pre-trained large language models in the prompt-specific setting, which can simultaneously achieve high accuracy and fairness. A similar conclusion can be drawn for the cross-prompt models. That is when pursuing generalizability, simple models based on well-investigated machine learning models such as SVM coupled with informative hand-crafted features might be preferable to complex models based on deep neural networks to achieve not only accurate but also fair essay evaluation. 

\section{Discussion and Conclusion}
To better support instructors and educators in selecting approximate AES models, we carefully selected nine representative AES approaches, covering both prompt-specific and cross-prompt categories. Subsequently, we evaluated the effectiveness of these methods on an open-sourced dataset with five demographic attributes using seven distinct metrics that account for both accuracy and fairness. Upon scrutinizing the results as detailed in Section Results, we derive the subsequent implications and acknowledge the limitations of our study.

\noindent \textbf{Implications.} Firstly, the results reveal a 9.00\% QWK gap, 8.71\% MAE gap, and 5.42\% PCC gap between the top-performing prompt-specific model (\texttt{BERT (3 Layers)}) and the best cross-prompt model (\texttt{SVM (Reduced)}). This suggests that choosing \texttt{SVM (Reduced)} could improve generalizability, although with some accuracy trade-offs. Secondly, \texttt{BERT (3 Layers)} excels in fairness (MAED of -0.005, just 0.001 apart from the best) and achieves the highest accuracy in prompt-specific settings, making it a strong recommendation for such settings. \texttt{CNN-LSTM-ATT} delivers top fairness (MAED of 0.004) and the second-best accuracy (2.7\% QWK decrease from the best) in prompt-specific settings, making it another strong recommendation.

\noindent \textbf{Limitations.} We acknowledged the following limitations of our study. Firstly, our experiments were restricted to a single dataset, underscoring the need to enhance the broader applicability of our findings through the inclusion of supplementary datasets in our evaluation process. Secondly, our analysis was predominantly centered around evaluating fairness, without providing definite solutions for addressing the identified fairness disparities. In future research, our emphasis will be on mitigating model unfairness while upholding an acceptable level of accuracy.

\section*{Acknowledgments}
This work is supported in part by National Key R\&D Program of China (2022YFC3303603), Australian Research Council (DP220101209), NSFC (62077028, 62377028), Key Laboratory of Smart Education of Guangdong Higher Education Institutes, Jinan University (2022LSYS003).

\bibliography{aaai24}

\section*{Appendix}
\begin{table}[!htb]
 \caption{The statistics of the dataset. The symbol $\sim$ denotes "not". \textbf{ED:} Economic Disadvantage. \textbf{ELL:} English Language Learner (Learning English as a second language).}
 \begin{center}
 \resizebox{0.48\textwidth}{!}{
\begin{tabular}{@{}lllllll@{}}
\toprule
\multicolumn{1}{c}{Prompt} & \multicolumn{1}{c}{\#Essays} & \multicolumn{1}{c}{Male}   & \multicolumn{1}{c}{$\sim$ED} & \multicolumn{1}{c}{$\sim$Disablity} & \multicolumn{1}{c}{$\sim$ELL} & \multicolumn{1}{c}{White}  \\ \midrule
\multicolumn{1}{c|}{1}     & \multicolumn{1}{c|}{2,157}    & \multicolumn{1}{c}{51.1\%} & \multicolumn{1}{c}{58.5\%}   & \multicolumn{1}{c}{85.0\%}          & \multicolumn{1}{c}{72.1\%}    & \multicolumn{1}{c}{32.4\%} \\ \midrule
\multicolumn{1}{c|}{2}     & \multicolumn{1}{c|}{2,132}    & \multicolumn{1}{c}{52.5\%} & \multicolumn{1}{c}{39.2\%}   & \multicolumn{1}{c}{69.7\%}          & \multicolumn{1}{c}{76.8\%}    & \multicolumn{1}{c}{39.0\%} \\ \midrule
\multicolumn{1}{c|}{3}     & \multicolumn{1}{c|}{1,856}    & \multicolumn{1}{c}{50.2\%} & \multicolumn{1}{c}{43.0\%}   & \multicolumn{1}{c}{70.5\%}          & \multicolumn{1}{c}{77.9\%}    & \multicolumn{1}{c}{43.7\%} \\ \midrule
\multicolumn{1}{c|}{4}     & \multicolumn{1}{c|}{1,809}    & \multicolumn{1}{c}{48.2\%} & \multicolumn{1}{c}{42.2\%}   & \multicolumn{1}{c}{95.5\%}          & \multicolumn{1}{c}{96.6\%}    & \multicolumn{1}{c}{44.3\%} \\ \midrule
\multicolumn{1}{c|}{5}     & \multicolumn{1}{c|}{1,750}    & \multicolumn{1}{c}{50.4\%} & \multicolumn{1}{c}{67.2\%}   & \multicolumn{1}{c}{91.70\%}         & \multicolumn{1}{c}{97.0\%}    & \multicolumn{1}{c}{40.9\%} \\ \midrule
\multicolumn{1}{c|}{6}     & \multicolumn{1}{c|}{1,670}    & \multicolumn{1}{c}{49.3\%} & \multicolumn{1}{c}{67.8\%}   & \multicolumn{1}{c}{84.0\%}          & \multicolumn{1}{c}{92.9\%}    & \multicolumn{1}{c}{45.9\%} \\ \midrule
\multicolumn{1}{c|}{7}     & \multicolumn{1}{c|}{1,633}    & \multicolumn{1}{c}{46.3\%} & \multicolumn{1}{c}{53.9\%}   & \multicolumn{1}{c}{91.9\%}          & \multicolumn{1}{c}{96.0\%}    & \multicolumn{1}{c}{51.3\%} \\ \midrule
\multicolumn{1}{c|}{8}     & \multicolumn{1}{c|}{1,606}    & \multicolumn{1}{c}{46.3\%} & \multicolumn{1}{c}{52.3\%}   & \multicolumn{1}{c}{93.3\%}          & \multicolumn{1}{c}{95.5\%}    & \multicolumn{1}{c}{51.5\%} \\ \midrule
\multicolumn{1}{c|}{9}     & \multicolumn{1}{c|}{1,572}    & \multicolumn{1}{c}{45.2\%} & \multicolumn{1}{c}{49.0\%}   & \multicolumn{1}{c}{96.6\%}          & \multicolumn{1}{c}{96.1\%}    & \multicolumn{1}{c}{48.3\%} \\ \midrule
\multicolumn{1}{c|}{10}    & \multicolumn{1}{c|}{1,552}    & \multicolumn{1}{c}{51.1\%} & \multicolumn{1}{c}{67.3\%}   & \multicolumn{1}{c}{89.0\%}          & \multicolumn{1}{c}{95.6\%}    & \multicolumn{1}{c}{49.7\%} \\ \midrule
\multicolumn{1}{c|}{11}    & \multicolumn{1}{c|}{1,521}    & \multicolumn{1}{c}{46.2\%} & \multicolumn{1}{c}{54.6\%}   & \multicolumn{1}{c}{93.6\%}          & \multicolumn{1}{c}{95.7\%}    & \multicolumn{1}{c}{51.0\%} \\ \midrule
\multicolumn{1}{c|}{12}    & \multicolumn{1}{c|}{1,368}    & \multicolumn{1}{c}{46.1\%} & \multicolumn{1}{c}{48.9\%}   & \multicolumn{1}{c}{94.7\%}          & \multicolumn{1}{c}{97.8\%}    & \multicolumn{1}{c}{53.3\%} \\ \midrule
\end{tabular}
}
 \end{center}
\end{table}

\begin{table*}[!htb]
\caption{Handcrafted Features} 
\resizebox{1\textwidth}{!}{
\begin{tabular}{@{}l|l|l@{}}
\toprule
Category                                  & Feature                                 & Description                                                          \\ \midrule
\multirow{4}{*}{Strongly Prompt-Dependent} &
  essay length &
  Text length by counting all tokens and sentences in an essay \\ \cmidrule(l){2-3} 
 &
  partition word n-gram &
  \begin{tabular}[c]{@{}l@{}}The 1,000 most frequent uni-, bi- and tri-grams in the partitions \\ (equally sized parts based on word counts) of the essay set\end{tabular} \\ \cmidrule(l){2-3} 
                                          & POS n-gram                              & The 1,000 most frequent POS uni-, bi- and tri-grams in the essay set \\ \cmidrule(l){2-3} 
                                          & word n-gram                             & The 1,000 most frequent uni-, bi- and tri-grams in the essay set     \\ \midrule
\multirow{12}{*}{Weakly Prompt-Dependent} & connectives                             & Occurrences of connectives                                           \\ \cmidrule(l){2-3} 
                                          & commas/quotations/exclamation           & Occurrences of commas/quotations/exclamation                         \\ \cmidrule(l){2-3} 
 &
  corpus similarity &
  \begin{tabular}[c]{@{}l@{}}Kullback–Leibler divergence between a neutral background \\ corpus (Brown corpus) and essay\end{tabular} \\ \cmidrule(l){2-3} 
                                          & formality                               & The relative ratio of POS-tags                                       \\ \cmidrule(l){2-3} 
                                          & grammar error                           & Occurrences of grammar error                                         \\ \cmidrule(l){2-3} 
                                          & readability                             & Flesch, Coleman-Liau, ARI, Kincaid, FOG, Lix, and SMOG               \\ \cmidrule(l){2-3} 
                                          & subordinate, causal \& temporal clauses & Occurrences of subordinate, causal and temporal clauses              \\ \cmidrule(l){2-3} 
                                          & topical overlap                         & N-gram overlap and redundancy between adjacent sentences             \\ \cmidrule(l){2-3} 
                                          & syntactic variation                     & The average depths of the parse trees                                \\ \cmidrule(l){2-3} 
 &
  type-token-ratio &
  \begin{tabular}[c]{@{}l@{}}The ratio obtained by dividing the the total number of different \\ words occurring by the total number of words\end{tabular} \\ \cmidrule(l){2-3} 
                                          & word frequency                          & The average word frequency based on the Brown corpus                 \\ \cmidrule(l){2-3} 
                                          & word/sentence length                    & The average sentence length in words and word length in characters   \\ \bottomrule
\end{tabular}
}
\end{table*}

\begin{table*}[!htb]
\caption{Experimental hyperparameters} 
\resizebox{1\textwidth}{!}{
\begin{tabular}{@{}l|l|l@{}}
\toprule
\textbf{Types of Models} &
  \textbf{Methods} &
  \textbf{Hyperparameters} \\ \midrule
\multirow{4}{*}{Prompt-Specific} &
  SKIPFLOW-LSTM &
  \begin{tabular}[c]{@{}l@{}}Epochs: 50. Vocabulary size: 4000. Batch size: 128. Hidden layer size: 50. Relevance width: 50.\\ Tensor slices: 5.  L: 500. Word embedding size: 300. \\ Loss function: Mean square error. Optimizer: Adam.\end{tabular} \\ \cmidrule(l){2-3} 
 &
  CNN-LSTM-ATT &
  \begin{tabular}[c]{@{}l@{}}Epochs: 50. Vocabulary size: 4000. Batch size: 10. Char embedding dim: 30. \\ Word embedding dim: 50. CNN window size: 5 Number of filters: 100. LSTM hidden units: 100. \\ Dropout: dropout rate: 0.5. Loss function: Mean square error. Optimizer: RMSprop.\end{tabular} \\ \cmidrule(l){2-3} 
 &
  R$^2$BERT &
  \begin{tabular}[c]{@{}l@{}}Epochs: 30. Batch size: 16. Document length: 512. BERT: bert-base-uncased. \\ Loss function: Combination of Mean square error and ListNet loss. Optimizer: Adam.\end{tabular} \\ \cmidrule(l){2-3} 
 &
  BERT (3 Layers) &
  \begin{tabular}[c]{@{}l@{}}Epochs: 30. Batch size: 16. Document length: 512. BERT: bert-base-uncased. \\ Loss function: Mean square error. Optimizer: Adam.\end{tabular} \\ \midrule
\multirow{3}{*}{Cross-Prompt} &
  Rank-SVM &
  Kernel: linear. C: 5. \\ \cmidrule(l){2-3} 
 &
  PAES &
  \begin{tabular}[c]{@{}l@{}}Epochs: 50. Vocabulary size: 4000. Batch size: 128. Char embedding dim: 30. \\ Word embedding dim: 50. CNN window size: 5 Number of filters: 100. LSTM hidden units: 100. \\ Dropout: dropout rate: 0.5. Loss function: Mean square error. Optimizer: RMSprop.\end{tabular} \\ \cmidrule(l){2-3} 
 &
  TDNN &
  \begin{tabular}[c]{@{}l@{}}Kernel: linear. C: 5.  Epochs: 50. Vocabulary size: 4000. Batch size: 16. \\ Sentence length: 40. Word embedding dim: 50. Hidden layer size: 50. \\ Dense layer size: 50.  Loss function: Mean square error. Optimizer: Adam.\end{tabular} \\ \bottomrule
\end{tabular}
}
\end{table*}

\begin{table*}[!htb]
\caption{The predictive \textbf{fairness} of the nine selected AES methods for \textbf{Gender}. The `ns' label indicates non-significant results (p $<$ 0.05). Lower values indicate a higher level of fairness of an AES model for all metrics.} 
\resizebox{1\textwidth}{!}{
\begin{tabular}{@{}l|l|cccc|cccc|cccc@{}}
\toprule
\multirow{2}{*}{\textbf{\begin{tabular}[c]{@{}l@{}}Types of Models\end{tabular}}} &
  \textbf{Metrics} &
  \multicolumn{1}{l}{\textbf{OSA}} &
  \multicolumn{1}{l}{\textbf{OSD}} &
  \multicolumn{1}{l}{\textbf{CSD}} &
  \multicolumn{1}{l|}{\textbf{MAED}} &
  \multicolumn{1}{l}{\textbf{OSA}} &
  \multicolumn{1}{l}{\textbf{OSD}} &
  \multicolumn{1}{l}{\textbf{CSD}} &
  \multicolumn{1}{l|}{\textbf{MAED}} &
  \multicolumn{1}{l}{\textbf{OSA}} &
  \multicolumn{1}{l}{\textbf{OSD}} &
  \multicolumn{1}{l}{\textbf{CSD}} &
  \multicolumn{1}{l}{\textbf{MAED}} \\ \cmidrule(l){2-14} 
 &
  \textbf{Methods} &
  \multicolumn{4}{c|}{\textbf{Prompt 1}} &
  \multicolumn{4}{c|}{\textbf{Prompt 2}} &
  \multicolumn{4}{c}{\textbf{Prompt 3}} \\ \midrule
\multicolumn{1}{c|}{\multirow{5}{*}{Prompt-Specific}} &
  SVM (Full) &
  ns &
  ns &
  ns &
  -0.011 &
  ns &
  ns &
  ns &
  -0.015 &
  ns &
  ns &
  ns &
  -0.032 \\
\multicolumn{1}{c|}{} &
  SKIPFLOW-LSTM &
  ns &
  ns &
  ns &
  -0.165 &
  ns &
  ns &
  ns &
  -0.197 &
  0.017 &
  0.018 &
  ns &
  -0.219 \\
\multicolumn{1}{c|}{} &
  CNN-LSTM-ATT &
  ns &
  ns &
  ns &
  -0.026 &
  ns &
  ns &
  ns &
  0.004 &
  ns &
  ns &
  ns &
  -0.024 \\
\multicolumn{1}{c|}{} &
  R$^2$BERT &
  ns &
  ns &
  ns &
  0.002 &
  ns &
  ns &
  ns &
  -0.032 &
  ns &
  ns &
  ns &
  -0.029 \\
\multicolumn{1}{c|}{} &
  BERT (3 Layers) &
  ns &
  ns &
  ns &
  -0.021 &
  ns &
  ns &
  ns &
  -0.026 &
  ns &
  ns &
  ns &
  -0.007 \\ \midrule
\multirow{4}{*}{Cross-Prompt} &
  SVM (Reduced) &
  ns &
  ns &
  ns &
  -0.014 &
  ns &
  ns &
  0.013 &
  -0.028 &
  ns &
  ns &
  0.009 &
  -0.011 \\
 &
  Rank-SVM &
  ns &
  ns &
  0.008 &
  -0.01 &
  ns &
  ns &
  ns &
  0.049 &
  ns &
  ns &
  ns &
  0.081 \\
 &
  PAES &
  ns &
  ns &
  ns &
  -0.013 &
  ns &
  ns &
  ns &
  0.002 &
  ns &
  ns &
  ns &
  -0.03 \\
 &
  TDNN &
  ns &
  ns &
  ns &
  0.031 &
  ns &
  ns &
  ns &
  0.017 &
  ns &
  ns &
  ns &
  0.041 \\ \midrule
\textbf{} &
   &
  \multicolumn{4}{c|}{\textbf{Prompt 4}} &
  \multicolumn{4}{c|}{\textbf{Prompt 5}} &
  \multicolumn{4}{c}{\textbf{Prompt 6}} \\ \midrule
\multicolumn{1}{c|}{\multirow{5}{*}{Prompt-Specific}} &
  SVM (Full) &
  ns &
  ns &
  ns &
  0.001 &
  ns &
  ns &
  ns &
  0.024 &
  ns &
  ns &
  ns &
  0.003 \\
\multicolumn{1}{c|}{} &
  SKIPFLOW-LSTM &
  ns &
  ns &
  ns &
  -0.112 &
  ns &
  ns &
  ns &
  -0.1 &
  0.017 &
  0.018 &
  ns &
  -0.189 \\
\multicolumn{1}{c|}{} &
  CNN-LSTM-ATT &
  ns &
  ns &
  ns &
  0.005 &
  ns &
  ns &
  ns &
  0.001 &
  ns &
  ns &
  ns &
  0.0 \\
\multicolumn{1}{c|}{} &
  R$^2$BERT &
  ns &
  ns &
  ns &
  -0.024 &
  ns &
  ns &
  ns &
  0.007 &
  ns &
  ns &
  ns &
  -0.006 \\
\multicolumn{1}{c|}{} &
  BERT (3 Layers) &
  ns &
  ns &
  ns &
  -0.009 &
  ns &
  ns &
  ns &
  0.024 &
  ns &
  ns &
  ns &
  0.015 \\ \midrule
\multirow{4}{*}{Cross-Prompt} &
  SVM (Reduced) &
  ns &
  ns &
  ns &
  -0.021 &
  ns &
  ns &
  ns &
  0.03 &
  ns &
  ns &
  ns &
  0.015 \\
 &
  Rank-SVM &
  ns &
  ns &
  ns &
  0.023 &
  ns &
  ns &
  ns &
  0.026 &
  ns &
  ns &
  ns &
  0.064 \\
 &
  PAES &
  ns &
  ns &
  ns &
  -0.036 &
  ns &
  ns &
  ns &
  0.039 &
  ns &
  ns &
  ns &
  0.077 \\
 &
  TDNN &
  ns &
  ns &
  ns &
  0.004 &
  ns &
  ns &
  ns &
  0.03 &
  ns &
  ns &
  ns &
  0.012 \\ \midrule
\textbf{} &
   &
  \multicolumn{4}{c|}{\textbf{Prompt 7}} &
  \multicolumn{4}{c|}{\textbf{Prompt 8}} &
  \multicolumn{4}{c}{\textbf{Prompt 9}} \\ \midrule
\multicolumn{1}{c|}{\multirow{5}{*}{Prompt-Specific}} &
  SVM (Full) &
  ns &
  ns &
  ns &
  -0.002 &
  ns &
  ns &
  ns &
  -0.015 &
  ns &
  ns &
  ns &
  0.009 \\
\multicolumn{1}{c|}{} &
  SKIPFLOW-LSTM &
  ns &
  ns &
  ns &
  -0.162 &
  ns &
  ns &
  ns &
  -0.167 &
  ns &
  ns &
  ns &
  -0.113 \\
\multicolumn{1}{c|}{} &
  CNN-LSTM-ATT &
  ns &
  ns &
  ns &
  0.012 &
  ns &
  ns &
  ns &
  -0.028 &
  ns &
  ns &
  ns &
  0.003 \\
\multicolumn{1}{c|}{} &
  R$^2$BERT &
  ns &
  ns &
  ns &
  -0.03 &
  ns &
  ns &
  ns &
  -0.027 &
  ns &
  ns &
  ns &
  -0.018 \\
\multicolumn{1}{c|}{} &
  BERT (3 Layers) &
  ns &
  ns &
  ns &
  -0.011 &
  ns &
  ns &
  ns &
  -0.021 &
  ns &
  ns &
  ns &
  -0.028 \\ \midrule
\multirow{4}{*}{Cross-Prompt} &
  SVM (Reduced) &
  ns &
  ns &
  0.011 &
  0.007 &
  ns &
  ns &
  0.01 &
  0.003 &
  ns &
  ns &
  ns &
  -0.014 \\
 &
  Rank-SVM &
  ns &
  ns &
  ns &
  0.078 &
  ns &
  ns &
  ns &
  0.045 &
  ns &
  ns &
  ns &
  0.022 \\
 &
  PAES &
  ns &
  ns &
  ns &
  0.021 &
  ns &
  ns &
  ns &
  -0.023 &
  ns &
  ns &
  ns &
  -0.025 \\
 &
  TDNN &
  ns &
  ns &
  ns &
  -0.071 &
  ns &
  ns &
  ns &
  -0.092 &
  ns &
  ns &
  ns &
  -0.037 \\ \midrule
\textbf{} &
  \textbf{} &
  \multicolumn{4}{c|}{\textbf{Prompt 10}} &
  \multicolumn{4}{c|}{\textbf{Prompt 11}} &
  \multicolumn{4}{c}{\textbf{Prompt 12}} \\ \midrule
\multicolumn{1}{c|}{\multirow{5}{*}{Prompt-Specific}} &
  SVM (Full) &
  ns &
  ns &
  ns &
  0.021 &
  ns &
  ns &
  ns &
  -0.012 &
  ns &
  ns &
  ns &
  -0.055 \\
\multicolumn{1}{c|}{} &
  SKIPFLOW-LSTM &
  ns &
  ns &
  ns &
  -0.164 &
  ns &
  ns &
  ns &
  -0.135 &
  ns &
  ns &
  ns &
  -0.078 \\
\multicolumn{1}{c|}{} &
  CNN-LSTM-ATT &
  ns &
  ns &
  ns &
  0.033 &
  ns &
  ns &
  ns &
  -0.035 &
  ns &
  ns &
  ns &
  -0.018 \\
\multicolumn{1}{c|}{} &
  R$^2$BERT &
  ns &
  ns &
  ns &
  0.03 &
  ns &
  ns &
  ns &
  -0.052 &
  ns &
  ns &
  ns &
  -0.038 \\
\multicolumn{1}{c|}{} &
  BERT (3 Layers) &
  ns &
  ns &
  ns &
  0.036 &
  ns &
  ns &
  ns &
  -0.048 &
  ns &
  ns &
  ns &
  -0.048 \\ \midrule
\multirow{4}{*}{Cross-Prompt} &
  SVM (Reduced) &
  ns &
  ns &
  ns &
  0.029 &
  ns &
  ns &
  ns &
  -0.035 &
  ns &
  ns &
  ns &
  -0.06 \\
 &
  Rank-SVM &
  ns &
  ns &
  ns &
  0.064 &
  ns &
  ns &
  ns &
  0.049 &
  ns &
  ns &
  ns &
  0.005 \\
 &
  PAES &
  ns &
  ns &
  ns &
  0.016 &
  ns &
  ns &
  ns &
  -0.052 &
  ns &
  ns &
  ns &
  -0.061 \\
 &
  TDNN &
  ns &
  ns &
  ns &
  0.059 &
  ns &
  0.02 &
  ns &
  0.069 &
  ns &
  ns &
  ns &
  -0.019 \\ \bottomrule
\end{tabular}
}
\end{table*}

\begin{table*}[!htb]
 \caption{The overall \textbf{fairness} performance of the nine selected AES methods for \textbf{Gender}. Cells in OSA, OSD, and CSD denote the number of prompts in which an AES method was diagnosed to have predictive bias, e.g., the number of cells with values other than `ns'. Cells in  MAED represent the average MAED of all prompts.}
 \begin{center}
 \resizebox{0.45\textwidth}{!}{
\begin{tabular}{@{}l|l|cccc@{}}
\toprule
Types of Models & Methods & \multicolumn{1}{l}{OSA} & \multicolumn{1}{l}{OSD} & \multicolumn{1}{l}{CSD} & \multicolumn{1}{l}{MAED} \\ \midrule
\multirow{5}{*}{Prompt-Specific} & SVM (Full)       & 0 & 0 & 0 & -0.004 \\ \cmidrule(l){2-6} 
                                 & SKIPFLOW-LSTM               & 2 & 2 & 0 & -0.159 \\ \cmidrule(l){2-6} 
                                 & CNN-LSTM-ATT                & 0 & 0 & 0 & -0.007 \\ \cmidrule(l){2-6} 
                                 & R$^2$BERT & 0 & 0 & 0 & -0.018 \\ \cmidrule(l){2-6} 
                                 & BERT (3 Layers)             & 0 & 0 & 0 & -0.010 \\ \midrule
\multirow{4}{*}{Cross-Prompt}    & SVM (Reduced)      & 0 & 0 & 4 & -0.005 \\ \cmidrule(l){2-6} 
                                 & Rank-SVM                    & 0 & 0 & 1 & 0.044  \\ \cmidrule(l){2-6} 
                                 & PAES                        & 0 & 0 & 0 & -0.002 \\ \cmidrule(l){2-6} 
                                 & TDNN                        & 0 & 1 & 0 & 0.005  \\ \bottomrule
\end{tabular}
}
 \end{center}
\end{table*}

\begin{table*}[!htb]
\caption{The predictive \textbf{fairness} of the nine selected AES methods for \textbf{Disability}. The `ns' label indicates non-significant results (p $<$ 0.05). Lower values indicate a higher level of fairness of an AES model for all metrics.} 
\resizebox{1\textwidth}{!}{
\begin{tabular}{@{}l|l|cccc|cccc|cccc@{}}
\toprule
\multirow{2}{*}{\textbf{\begin{tabular}[c]{@{}l@{}}Types of Models\end{tabular}}} &
  \textbf{Metrics} &
  \multicolumn{1}{l}{\textbf{OSA}} &
  \multicolumn{1}{l}{\textbf{OSD}} &
  \multicolumn{1}{l}{\textbf{CSD}} &
  \multicolumn{1}{l|}{\textbf{MAED}} &
  \multicolumn{1}{l}{\textbf{OSA}} &
  \multicolumn{1}{l}{\textbf{OSD}} &
  \multicolumn{1}{l}{\textbf{CSD}} &
  \multicolumn{1}{l|}{\textbf{MAED}} &
  \multicolumn{1}{l}{\textbf{OSA}} &
  \multicolumn{1}{l}{\textbf{OSD}} &
  \multicolumn{1}{l}{\textbf{CSD}} &
  \multicolumn{1}{l}{\textbf{MAED}} \\ \cmidrule(l){2-14} 
 &
  \textbf{Methods} &
  \multicolumn{4}{c|}{\textbf{Prompt 1}} &
  \multicolumn{4}{c|}{\textbf{Prompt 2}} &
  \multicolumn{4}{c}{\textbf{Prompt 3}} \\ \midrule
\multicolumn{1}{c|}{\multirow{5}{*}{Prompt-Specific}} &
  SVM (Full) &
  ns &
  ns &
  ns &
  -0.006 &
  ns &
  0.024 &
  ns &
  -0.028 &
  ns &
  ns &
  0.01 &
  -0.031 \\
\multicolumn{1}{c|}{} &
  SKIPFLOW-LSTM &
  0.032 &
  0.031 &
  ns &
  -0.375 &
  0.081 &
  0.092 &
  ns &
  -0.525 &
  0.083 &
  0.094 &
  ns &
  -0.561 \\
\multicolumn{1}{c|}{} &
  CNN-LSTM-ATT &
  ns &
  ns &
  ns &
  0.02 &
  ns &
  ns &
  0.007 &
  -0.014 &
  ns &
  ns &
  ns &
  -0.015 \\
\multicolumn{1}{c|}{} &
  R$^2$BERT &
  ns &
  ns &
  ns &
  -0.036 &
  ns &
  ns &
  ns &
  -0.037 &
  ns &
  ns &
  0.022 &
  -0.035 \\
\multicolumn{1}{c|}{} &
  BERT (3 Layers) &
  ns &
  ns &
  ns &
  -0.006 &
  ns &
  ns &
  0.013 &
  -0.023 &
  ns &
  ns &
  0.017 &
  -0.024 \\ \midrule
\multirow{4}{*}{Cross-Prompt} &
  SVM (Reduced) &
  ns &
  ns &
  ns &
  -0.005 &
  ns &
  ns &
  ns &
  0.042 &
  ns &
  ns &
  ns &
  -0.019 \\
 &
  Rank-SVM &
  ns &
  ns &
  ns &
  -0.002 &
  ns &
  ns &
  ns &
  0.145 &
  0.014 &
  0.035 &
  ns &
  0.175 \\
 &
  PAES &
  ns &
  ns &
  ns &
  0.008 &
  ns &
  ns &
  ns &
  0.033 &
  ns &
  ns &
  0.021 &
  -0.019 \\
 &
  TDNN &
  ns &
  ns &
  ns &
  0.028 &
  ns &
  0.031 &
  ns &
  0.056 &
  ns &
  0.047 &
  ns &
  0.093 \\ \midrule
\textbf{} &
   &
  \multicolumn{4}{c|}{\textbf{Prompt 4}} &
  \multicolumn{4}{c|}{\textbf{Prompt 5}} &
  \multicolumn{4}{c}{\textbf{Prompt 6}} \\ \midrule
\multicolumn{1}{c|}{\multirow{5}{*}{Prompt-Specific}} &
  SVM (Full) &
  ns &
  ns &
  ns &
  0.014 &
  ns &
  ns &
  ns &
  0.042 &
  ns &
  ns &
  ns &
  0.077 \\
\multicolumn{1}{c|}{} &
  SKIPFLOW-LSTM &
  ns &
  ns &
  ns &
  -0.316 &
  ns &
  ns &
  ns &
  -0.255 &
  0.039 &
  0.044 &
  ns &
  -0.401 \\
\multicolumn{1}{c|}{} &
  CNN-LSTM-ATT &
  ns &
  ns &
  ns &
  0.015 &
  ns &
  ns &
  ns &
  0.043 &
  ns &
  ns &
  ns &
  0.053 \\
\multicolumn{1}{c|}{} &
  R$^2$BERT &
  ns &
  ns &
  ns &
  0.006 &
  ns &
  ns &
  ns &
  0.003 &
  ns &
  ns &
  ns &
  0.048 \\
\multicolumn{1}{c|}{} &
  BERT (3 Layers) &
  ns &
  ns &
  ns &
  0.037 &
  ns &
  ns &
  ns &
  0.026 &
  ns &
  ns &
  ns &
  0.035 \\ \midrule
\multirow{4}{*}{Cross-Prompt} &
  SVM (Reduced) &
  ns &
  ns &
  ns &
  0.007 &
  ns &
  ns &
  ns &
  0.012 &
  ns &
  ns &
  ns &
  0.027 \\
 &
  Rank-SVM &
  ns &
  ns &
  ns &
  0.086 &
  ns &
  ns &
  ns &
  0.04 &
  ns &
  ns &
  ns &
  0.094 \\
 &
  PAES &
  ns &
  ns &
  ns &
  -0.019 &
  ns &
  ns &
  ns &
  0.025 &
  ns &
  ns &
  ns &
  0.098 \\
 &
  TDNN &
  ns &
  ns &
  ns &
  0.05 &
  ns &
  ns &
  ns &
  0.093 &
  ns &
  ns &
  ns &
  0.025 \\ \midrule
\textbf{} &
   &
  \multicolumn{4}{c|}{\textbf{Prompt 7}} &
  \multicolumn{4}{c|}{\textbf{Prompt 8}} &
  \multicolumn{4}{c}{\textbf{Prompt 9}} \\ \midrule
\multicolumn{1}{c|}{\multirow{5}{*}{Prompt-Specific}} &
  SVM (Full) &
  ns &
  ns &
  ns &
  0.004 &
  ns &
  ns &
  ns &
  0.124 &
  ns &
  ns &
  ns &
  0.09 \\
\multicolumn{1}{c|}{} &
  SKIPFLOW-LSTM &
  ns &
  ns &
  ns &
  -0.253 &
  ns &
  ns &
  ns &
  -0.22 &
  ns &
  ns &
  ns &
  -0.298 \\
\multicolumn{1}{c|}{} &
  CNN-LSTM-ATT &
  ns &
  ns &
  ns &
  -0.007 &
  ns &
  ns &
  ns &
  0.048 &
  ns &
  ns &
  ns &
  0.085 \\
\multicolumn{1}{c|}{} &
  R$^2$BERT &
  ns &
  ns &
  ns &
  -0.028 &
  ns &
  ns &
  ns &
  0.027 &
  ns &
  ns &
  ns &
  0.073 \\
\multicolumn{1}{c|}{} &
  BERT (3 Layers) &
  ns &
  ns &
  ns &
  0.007 &
  ns &
  ns &
  ns &
  0.022 &
  ns &
  ns &
  ns &
  0.068 \\ \midrule
\multirow{4}{*}{Cross-Prompt} &
  SVM (Reduced) &
  ns &
  ns &
  ns &
  -0.068 &
  ns &
  ns &
  ns &
  0.055 &
  ns &
  ns &
  ns &
  -0.01 \\
 &
  Rank-SVM &
  ns &
  ns &
  ns &
  0.067 &
  ns &
  ns &
  ns &
  0.0 &
  ns &
  ns &
  ns &
  0.091 \\
 &
  PAES &
  ns &
  ns &
  ns &
  -0.008 &
  ns &
  ns &
  ns &
  -0.006 &
  ns &
  ns &
  ns &
  -0.094 \\
 &
  TDNN &
  ns &
  ns &
  ns &
  -0.26 &
  ns &
  ns &
  ns &
  -0.144 &
  ns &
  ns &
  ns &
  0.003 \\ \midrule
\textbf{} &
  \textbf{} &
  \multicolumn{4}{c|}{\textbf{Prompt 10}} &
  \multicolumn{4}{c|}{\textbf{Prompt 11}} &
  \multicolumn{4}{c}{\textbf{Prompt 12}} \\ \midrule
\multicolumn{1}{c|}{\multirow{5}{*}{Prompt-Specific}} &
  SVM (Full) &
  ns &
  ns &
  ns &
  0.069 &
  ns &
  ns &
  ns &
  0.031 &
  ns &
  ns &
  ns &
  -0.064 \\
\multicolumn{1}{c|}{} &
  SKIPFLOW-LSTM &
  ns &
  ns &
  ns &
  -0.212 &
  ns &
  ns &
  ns &
  -0.244 &
  ns &
  ns &
  ns &
  -0.195 \\
\multicolumn{1}{c|}{} &
  CNN-LSTM-ATT &
  ns &
  ns &
  ns &
  0.085 &
  ns &
  ns &
  ns &
  -0.076 &
  ns &
  ns &
  ns &
  0.007 \\
\multicolumn{1}{c|}{} &
  R$^2$BERT &
  ns &
  ns &
  ns &
  0.083 &
  ns &
  ns &
  ns &
  -0.075 &
  ns &
  ns &
  ns &
  -0.036 \\
\multicolumn{1}{c|}{} &
  BERT (3 Layers) &
  ns &
  ns &
  ns &
  0.024 &
  ns &
  ns &
  ns &
  -0.08 &
  ns &
  ns &
  ns &
  -0.035 \\ \midrule
\multirow{4}{*}{Cross-Prompt} &
  SVM (Reduced) &
  ns &
  ns &
  ns &
  0.058 &
  ns &
  ns &
  ns &
  -0.034 &
  ns &
  ns &
  ns &
  -0.082 \\
 &
  Rank-SVM &
  ns &
  ns &
  ns &
  0.093 &
  ns &
  ns &
  ns &
  0.245 &
  ns &
  ns &
  ns &
  0.108 \\
 &
  PAES &
  ns &
  ns &
  ns &
  0.038 &
  ns &
  ns &
  ns &
  -0.081 &
  ns &
  ns &
  ns &
  -0.085 \\
 &
  TDNN &
  ns &
  ns &
  ns &
  0.071 &
  ns &
  0.027 &
  ns &
  0.222 &
  ns &
  ns &
  ns &
  0.045 \\ \bottomrule
\end{tabular}
}
\end{table*}

\begin{table*}[!htb]
 \caption{The overall \textbf{fairness} performance of the nine selected AES methods for \textbf{Disability}. Cells in OSA, OSD, and CSD denote the number of prompts in which an AES method was diagnosed to have predictive bias, e.g., the number of cells with values other than `ns'. Cells in  MAED represent the average MAED of all prompts.} 
 \begin{center}
 \resizebox{0.45\textwidth}{!}{
\begin{tabular}{@{}l|l|cccc@{}}
\toprule
Types of Models & Methods & \multicolumn{1}{l}{OSA} & \multicolumn{1}{l}{OSD} & \multicolumn{1}{l}{CSD} & \multicolumn{1}{l}{MAED} \\ \midrule
\multirow{5}{*}{Prompt-Specific} & SVM (Full)       & 0 & 1 & 1 & 0.027  \\ \cmidrule(l){2-6} 
                                 & SKIPFLOW-LSTM               & 4 & 4 & 0 & -0.321 \\ \cmidrule(l){2-6} 
                                 & CNN-LSTM-ATT                & 0 & 0 & 1 & 0.020  \\ \cmidrule(l){2-6} 
                                 & R$^2$BERT & 0 & 0 & 1 & -0.001 \\ \cmidrule(l){2-6} 
                                 & BERT (3 Layers)             & 0 & 0 & 2 & 0.004  \\ \midrule
\multirow{4}{*}{Cross-Prompt}    & SVM (Reduced)      & 0 & 0 & 0 & -0.001 \\ \cmidrule(l){2-6} 
                                 & Rank-SVM                    & 1 & 1 & 0 & 0.095  \\ \cmidrule(l){2-6} 
                                 & PAES                        & 0 & 0 & 1 & -0.009 \\ \cmidrule(l){2-6} 
                                 & TDNN                        & 0 & 3 & 0 & 0.024  \\ \bottomrule
\end{tabular}
}
 \end{center}
\end{table*}

\begin{table*}[!htb]
\caption{The predictive \textbf{fairness} of the nine selected AES methods for \textbf{English Language Learner Status}. The `ns' label indicates non-significant results (p $<$ 0.05). Lower values indicate a higher level of fairness of an AES model for all metrics.} 
\resizebox{1\textwidth}{!}{
\begin{tabular}{@{}l|l|cccc|cccc|cccc@{}}
\toprule
\multirow{2}{*}{\textbf{\begin{tabular}[c]{@{}l@{}}Types of Models\end{tabular}}} &
  \textbf{Metrics} &
  \multicolumn{1}{l}{\textbf{OSA}} &
  \multicolumn{1}{l}{\textbf{OSD}} &
  \multicolumn{1}{l}{\textbf{CSD}} &
  \multicolumn{1}{l|}{\textbf{MAED}} &
  \multicolumn{1}{l}{\textbf{OSA}} &
  \multicolumn{1}{l}{\textbf{OSD}} &
  \multicolumn{1}{l}{\textbf{CSD}} &
  \multicolumn{1}{l|}{\textbf{MAED}} &
  \multicolumn{1}{l}{\textbf{OSA}} &
  \multicolumn{1}{l}{\textbf{OSD}} &
  \multicolumn{1}{l}{\textbf{CSD}} &
  \multicolumn{1}{l}{\textbf{MAED}} \\ \cmidrule(l){2-14} 
 &
  \textbf{Methods} &
  \multicolumn{4}{c|}{\textbf{Prompt 1}} &
  \multicolumn{4}{c|}{\textbf{Prompt 2}} &
  \multicolumn{4}{c}{\textbf{Prompt 3}} \\ \midrule
\multicolumn{1}{c|}{\multirow{5}{*}{Prompt-Specific}} &
  SVM (Full) &
  0.037 &
  0.05 &
  0.014 &
  0.139 &
  ns &
  ns &
  ns &
  -0.014 &
  ns &
  0.031 &
  ns &
  0.07 \\
\multicolumn{1}{c|}{} &
  SKIPFLOW-LSTM &
  0.306 &
  0.323 &
  0.039 &
  -0.948 &
  0.039 &
  0.043 &
  ns &
  -0.4 &
  0.053 &
  0.054 &
  ns &
  -0.464 \\
\multicolumn{1}{c|}{} &
  CNN-LSTM-ATT &
  ns &
  ns &
  0.064 &
  0.083 &
  ns &
  ns &
  ns &
  -0.038 &
  ns &
  ns &
  ns &
  0.063 \\
\multicolumn{1}{c|}{} &
  R$^2$BERT &
  ns &
  ns &
  0.075 &
  -0.026 &
  ns &
  ns &
  ns &
  -0.069 &
  ns &
  ns &
  ns &
  0.053 \\
\multicolumn{1}{c|}{} &
  BERT (3 Layers) &
  ns &
  ns &
  0.064 &
  0.094 &
  ns &
  ns &
  ns &
  -0.009 &
  ns &
  ns &
  ns &
  0.044 \\ \midrule
\multirow{4}{*}{Cross-Prompt} &
  SVM (Reduced) &
  0.018 &
  0.038 &
  ns &
  0.07 &
  ns &
  0.021 &
  ns &
  0.091 &
  0.014 &
  0.03 &
  ns &
  0.146 \\
 &
  Rank-SVM &
  ns &
  0.067 &
  ns &
  0.027 &
  ns &
  ns &
  ns &
  0.15 &
  0.031 &
  0.05 &
  ns &
  0.323 \\
 &
  PAES &
  0.047 &
  ns &
  ns &
  0.193 &
  ns &
  ns &
  ns &
  0.04 &
  ns &
  ns &
  ns &
  0.165 \\
 &
  TDNN &
  ns &
  ns &
  0.026 &
  0.178 &
  ns &
  ns &
  ns &
  0.073 &
  ns &
  0.043 &
  ns &
  0.154 \\ \midrule
\textbf{} &
   &
  \multicolumn{4}{c|}{\textbf{Prompt 4}} &
  \multicolumn{4}{c|}{\textbf{Prompt 5}} &
  \multicolumn{4}{c}{\textbf{Prompt 6}} \\ \midrule
\multicolumn{1}{c|}{\multirow{5}{*}{Prompt-Specific}} &
  SVM (Full) &
  ns &
  ns &
  ns &
  0.04 &
  ns &
  ns &
  ns &
  0.275 &
  ns &
  ns &
  ns &
  0.105 \\
\multicolumn{1}{c|}{} &
  SKIPFLOW-LSTM &
  ns &
  ns &
  ns &
  -0.19 &
  0.02 &
  0.021 &
  ns &
  -0.527 &
  0.039 &
  0.045 &
  ns &
  -0.582 \\
\multicolumn{1}{c|}{} &
  CNN-LSTM-ATT &
  ns &
  ns &
  ns &
  0.035 &
  ns &
  ns &
  ns &
  0.099 &
  ns &
  ns &
  ns &
  0.047 \\
\multicolumn{1}{c|}{} &
  R$^2$BERT &
  ns &
  ns &
  ns &
  -0.006 &
  ns &
  ns &
  ns &
  0.05 &
  ns &
  ns &
  ns &
  -0.001 \\
\multicolumn{1}{c|}{} &
  BERT (3 Layers) &
  ns &
  ns &
  ns &
  0.006 &
  ns &
  ns &
  ns &
  0.107 &
  ns &
  ns &
  ns &
  0.028 \\ \midrule
\multirow{4}{*}{Cross-Prompt} &
  SVM (Reduced) &
  ns &
  ns &
  ns &
  0.024 &
  ns &
  ns &
  ns &
  0.247 &
  ns &
  ns &
  ns &
  0.092 \\
 &
  Rank-SVM &
  ns &
  ns &
  ns &
  0.092 &
  ns &
  0.03 &
  ns &
  0.268 &
  ns &
  0.018 &
  ns &
  0.238 \\
 &
  PAES &
  ns &
  ns &
  ns &
  -0.119 &
  ns &
  ns &
  ns &
  0.287 &
  ns &
  ns &
  ns &
  0.217 \\
 &
  TDNN &
  ns &
  ns &
  ns &
  0.022 &
  ns &
  ns &
  ns &
  0.233 &
  ns &
  0.015 &
  ns &
  0.059 \\ \midrule
\textbf{} &
   &
  \multicolumn{4}{c|}{\textbf{Prompt 7}} &
  \multicolumn{4}{c|}{\textbf{Prompt 8}} &
  \multicolumn{4}{c}{\textbf{Prompt 9}} \\ \midrule
\multicolumn{1}{c|}{\multirow{5}{*}{Prompt-Specific}} &
  SVM (Full) &
  ns &
  ns &
  ns &
  -0.075 &
  ns &
  ns &
  ns &
  0.082 &
  ns &
  ns &
  ns &
  0.018 \\
\multicolumn{1}{c|}{} &
  SKIPFLOW-LSTM &
  0.012 &
  0.014 &
  ns &
  -0.367 &
  ns &
  ns &
  ns &
  -0.268 &
  ns &
  ns &
  ns &
  -0.374 \\
\multicolumn{1}{c|}{} &
  CNN-LSTM-ATT &
  ns &
  ns &
  ns &
  -0.058 &
  ns &
  ns &
  ns &
  0.021 &
  ns &
  ns &
  ns &
  0.007 \\
\multicolumn{1}{c|}{} &
  R$^2$BERT &
  ns &
  ns &
  ns &
  -0.134 &
  ns &
  ns &
  ns &
  -0.023 &
  ns &
  ns &
  ns &
  -0.069 \\
\multicolumn{1}{c|}{} &
  BERT (3 Layers) &
  ns &
  ns &
  ns &
  -0.073 &
  ns &
  ns &
  ns &
  0.023 &
  ns &
  ns &
  ns &
  -0.011 \\ \midrule
\multirow{4}{*}{Cross-Prompt} &
  SVM (Reduced) &
  ns &
  ns &
  ns &
  -0.102 &
  ns &
  ns &
  ns &
  0.017 &
  ns &
  ns &
  ns &
  -0.038 \\
 &
  Rank-SVM &
  ns &
  ns &
  ns &
  -0.095 &
  ns &
  ns &
  ns &
  0.128 &
  ns &
  ns &
  ns &
  0.008 \\
 &
  PAES &
  ns &
  ns &
  ns &
  -0.036 &
  ns &
  ns &
  ns &
  -0.036 &
  ns &
  ns &
  ns &
  -0.121 \\
 &
  TDNN &
  ns &
  ns &
  ns &
  -0.367 &
  ns &
  0.018 &
  ns &
  -0.17 &
  ns &
  ns &
  ns &
  -0.184 \\ \midrule
\textbf{} &
  \textbf{} &
  \multicolumn{4}{c|}{\textbf{Prompt 10}} &
  \multicolumn{4}{c|}{\textbf{Prompt 11}} &
  \multicolumn{4}{c}{\textbf{Prompt 12}} \\ \midrule
\multicolumn{1}{c|}{\multirow{5}{*}{Prompt-Specific}} &
  SVM (Full) &
  ns &
  ns &
  ns &
  0.139 &
  ns &
  ns &
  ns &
  0.065 &
  ns &
  ns &
  ns &
  0.126 \\
\multicolumn{1}{c|}{} &
  SKIPFLOW-LSTM &
  0.013 &
  0.014 &
  ns &
  -0.358 &
  ns &
  ns &
  ns &
  -0.233 &
  ns &
  ns &
  ns &
  -0.418 \\
\multicolumn{1}{c|}{} &
  CNN-LSTM-ATT &
  ns &
  ns &
  ns &
  0.035 &
  ns &
  ns &
  ns &
  0.01 &
  ns &
  ns &
  ns &
  -0.119 \\
\multicolumn{1}{c|}{} &
  R$^2$BERT &
  ns &
  ns &
  ns &
  0.063 &
  ns &
  ns &
  ns &
  -0.1 &
  ns &
  ns &
  ns &
  0.021 \\
\multicolumn{1}{c|}{} &
  BERT (3 Layers) &
  ns &
  ns &
  ns &
  0.107 &
  ns &
  ns &
  ns &
  -0.065 &
  ns &
  ns &
  ns &
  -0.02 \\ \midrule
\multirow{4}{*}{Cross-Prompt} &
  SVM (Reduced) &
  ns &
  ns &
  ns &
  0.029 &
  ns &
  ns &
  ns &
  0.073 &
  ns &
  ns &
  ns &
  0.075 \\
 &
  Rank-SVM &
  ns &
  ns &
  ns &
  0.003 &
  ns &
  ns &
  ns &
  0.228 &
  ns &
  ns &
  ns &
  0.239 \\
 &
  PAES &
  ns &
  ns &
  ns &
  0.124 &
  ns &
  ns &
  ns &
  -0.084 &
  ns &
  ns &
  ns &
  0.119 \\
 &
  TDNN &
  ns &
  ns &
  ns &
  -0.052 &
  ns &
  ns &
  ns &
  0.304 &
  ns &
  ns &
  ns &
  0.152 \\ \bottomrule
\end{tabular}
}
\end{table*}

\begin{table*}[!htb]
 \caption{The overall \textbf{fairness} performance of the nine selected AES methods for \textbf{English Language Learner Status}. Cells in OSA, OSD, and CSD denote the number of prompts in which an AES method was diagnosed to have predictive bias, e.g., the number of cells with values other than `ns'. Cells in  MAED represent the average MAED of all prompts.} 
 \begin{center}
 \resizebox{0.45\textwidth}{!}{
\begin{tabular}{@{}l|l|cccc@{}}
\toprule
Types of Models & Methods & \multicolumn{1}{l}{OSA} & \multicolumn{1}{l}{OSD} & \multicolumn{1}{l}{CSD} & \multicolumn{1}{l}{MAED} \\ \midrule
\multirow{5}{*}{Prompt-Specific} & SVM (Full)       & 1 & 2 & 1 & 0.081  \\ \cmidrule(l){2-6} 
                                 & SKIPFLOW-LSTM               & 7 & 7 & 1 & -0.427 \\ \cmidrule(l){2-6} 
                                 & CNN-LSTM-ATT                & 0 & 0 & 1 & 0.015  \\ \cmidrule(l){2-6} 
                                 & R$^2$BERT & 0 & 0 & 1 & -0.020 \\ \cmidrule(l){2-6} 
                                 & BERT (3 Layers)             & 0 & 0 & 1 & 0.019  \\ \midrule
\multirow{4}{*}{Cross-Prompt}    & SVM (Reduced)      & 2 & 3 & 0 & 0.060  \\ \cmidrule(l){2-6} 
                                 & Rank-SVM                    & 1 & 4 & 0 & 0.134  \\ \cmidrule(l){2-6} 
                                 & PAES                        & 1 & 0 & 0 & 0.062  \\ \cmidrule(l){2-6} 
                                 & TDNN                        & 0 & 3 & 1 & 0.034  \\ \bottomrule
\end{tabular}

}
 \end{center}
\end{table*}

\begin{table*}[!htb]
\caption{The predictive \textbf{fairness} of the nine selected AES methods for \textbf{Race}. The `ns' label indicates non-significant results (p $<$ 0.05). Lower values indicate a higher level of fairness of an AES model for all metrics.} 
\resizebox{1\textwidth}{!}{
\begin{tabular}{@{}l|l|cccc|cccc|cccc@{}}
\toprule
\multirow{2}{*}{\textbf{\begin{tabular}[c]{@{}l@{}}Types of Models\end{tabular}}} &
  \textbf{Metrics} &
  \multicolumn{1}{l}{\textbf{OSA}} &
  \multicolumn{1}{l}{\textbf{OSD}} &
  \multicolumn{1}{l}{\textbf{CSD}} &
  \multicolumn{1}{l|}{\textbf{MAED}} &
  \multicolumn{1}{l}{\textbf{OSA}} &
  \multicolumn{1}{l}{\textbf{OSD}} &
  \multicolumn{1}{l}{\textbf{CSD}} &
  \multicolumn{1}{l|}{\textbf{MAED}} &
  \multicolumn{1}{l}{\textbf{OSA}} &
  \multicolumn{1}{l}{\textbf{OSD}} &
  \multicolumn{1}{l}{\textbf{CSD}} &
  \multicolumn{1}{l}{\textbf{MAED}} \\ \cmidrule(l){2-14} 
 &
  \textbf{Methods} &
  \multicolumn{4}{c|}{\textbf{Prompt 1}} &
  \multicolumn{4}{c|}{\textbf{Prompt 2}} &
  \multicolumn{4}{c}{\textbf{Prompt 3}} \\ \midrule
\multicolumn{1}{c|}{\multirow{5}{*}{Prompt-Specific}} &
  SVM (Full) &
  ns &
  ns &
  ns &
  -0.068 &
  ns &
  ns &
  ns &
  0.058 &
  ns &
  0.028 &
  ns &
  -0.01 \\
\multicolumn{1}{c|}{} &
  SKIPFLOW-LSTM &
  0.076 &
  0.077 &
  ns &
  0.447 &
  0.044 &
  0.046 &
  ns &
  0.35 &
  0.03 &
  0.03 &
  ns &
  0.299 \\
\multicolumn{1}{c|}{} &
  CNN-LSTM-ATT &
  ns &
  ns &
  ns &
  -0.037 &
  ns &
  ns &
  ns &
  0.039 &
  ns &
  ns &
  ns &
  -0.029 \\
\multicolumn{1}{c|}{} &
  R$^2$BERT &
  ns &
  ns &
  ns &
  -0.0 &
  ns &
  ns &
  ns &
  0.083 &
  ns &
  ns &
  ns &
  -0.026 \\
\multicolumn{1}{c|}{} &
  BERT (3 Layers) &
  ns &
  ns &
  ns &
  -0.055 &
  ns &
  ns &
  ns &
  0.033 &
  ns &
  ns &
  ns &
  -0.015 \\ \midrule
\multirow{4}{*}{Cross-Prompt} &
  SVM (Reduced) &
  ns &
  ns &
  ns &
  -0.035 &
  ns &
  ns &
  ns &
  -0.025 &
  ns &
  ns &
  ns &
  -0.053 \\
 &
  Rank-SVM &
  ns &
  0.013 &
  ns &
  0.001 &
  ns &
  0.028 &
  ns &
  -0.123 &
  0.014 &
  0.034 &
  ns &
  -0.168 \\
 &
  PAES &
  ns &
  ns &
  ns &
  -0.091 &
  ns &
  ns &
  ns &
  -0.008 &
  ns &
  ns &
  ns &
  -0.077 \\
 &
  TDNN &
  ns &
  ns &
  ns &
  -0.113 &
  ns &
  0.024 &
  ns &
  -0.082 &
  ns &
  0.039 &
  ns &
  -0.065 \\ \midrule
\textbf{} &
   &
  \multicolumn{4}{c|}{\textbf{Prompt 4}} &
  \multicolumn{4}{c|}{\textbf{Prompt 5}} &
  \multicolumn{4}{c}{\textbf{Prompt 6}} \\ \midrule
\multicolumn{1}{c|}{\multirow{5}{*}{Prompt-Specific}} &
  SVM (Full) &
  ns &
  ns &
  ns &
  -0.012 &
  ns &
  ns &
  ns &
  -0.0 &
  ns &
  ns &
  ns &
  -0.009 \\
\multicolumn{1}{c|}{} &
  SKIPFLOW-LSTM &
  ns &
  ns &
  ns &
  0.119 &
  ns &
  ns &
  ns &
  0.138 &
  ns &
  ns &
  ns &
  0.152 \\
\multicolumn{1}{c|}{} &
  CNN-LSTM-ATT &
  ns &
  ns &
  ns &
  -0.023 &
  ns &
  ns &
  ns &
  0.008 &
  ns &
  ns &
  ns &
  0.012 \\
\multicolumn{1}{c|}{} &
  R$^2$BERT &
  ns &
  ns &
  ns &
  -0.004 &
  ns &
  ns &
  ns &
  -0.011 &
  ns &
  ns &
  ns &
  0.004 \\
\multicolumn{1}{c|}{} &
  BERT (3 Layers) &
  ns &
  ns &
  ns &
  -0.006 &
  ns &
  ns &
  ns &
  -0.002 &
  ns &
  ns &
  ns &
  0.01 \\ \midrule
\multirow{4}{*}{Cross-Prompt} &
  SVM (Reduced) &
  ns &
  ns &
  ns &
  -0.019 &
  ns &
  ns &
  ns &
  0.002 &
  ns &
  ns &
  ns &
  -0.003 \\
 &
  Rank-SVM &
  ns &
  ns &
  0.005 &
  -0.041 &
  ns &
  ns &
  ns &
  -0.007 &
  ns &
  ns &
  ns &
  -0.022 \\
 &
  PAES &
  ns &
  ns &
  0.017 &
  0.015 &
  ns &
  ns &
  ns &
  -0.036 &
  ns &
  ns &
  ns &
  -0.055 \\
 &
  TDNN &
  ns &
  ns &
  ns &
  -0.019 &
  ns &
  ns &
  ns &
  -0.005 &
  ns &
  ns &
  ns &
  0.013 \\ \midrule
\textbf{} &
   &
  \multicolumn{4}{c|}{\textbf{Prompt 7}} &
  \multicolumn{4}{c|}{\textbf{Prompt 8}} &
  \multicolumn{4}{c}{\textbf{Prompt 9}} \\ \midrule
\multicolumn{1}{c|}{\multirow{5}{*}{Prompt-Specific}} &
  SVM (Full) &
  ns &
  ns &
  ns &
  0.019 &
  ns &
  ns &
  ns &
  0.029 &
  ns &
  ns &
  ns &
  0.043 \\
\multicolumn{1}{c|}{} &
  SKIPFLOW-LSTM &
  ns &
  ns &
  ns &
  0.173 &
  0.011 &
  0.012 &
  ns &
  0.146 &
  0.027 &
  0.028 &
  ns &
  0.228 \\
\multicolumn{1}{c|}{} &
  CNN-LSTM-ATT &
  ns &
  ns &
  ns &
  0.003 &
  ns &
  ns &
  ns &
  0.006 &
  ns &
  ns &
  ns &
  0.033 \\
\multicolumn{1}{c|}{} &
  R$^2$BERT &
  ns &
  ns &
  ns &
  0.023 &
  ns &
  ns &
  ns &
  0.018 &
  ns &
  ns &
  ns &
  0.061 \\
\multicolumn{1}{c|}{} &
  BERT (3 Layers) &
  ns &
  ns &
  ns &
  0.021 &
  ns &
  ns &
  ns &
  0.027 &
  ns &
  ns &
  ns &
  0.032 \\ \midrule
\multirow{4}{*}{Cross-Prompt} &
  SVM (Reduced) &
  ns &
  ns &
  ns &
  0.013 &
  ns &
  ns &
  ns &
  0.011 &
  ns &
  ns &
  ns &
  0.061 \\
 &
  Rank-SVM &
  ns &
  ns &
  ns &
  -0.039 &
  ns &
  ns &
  ns &
  -0.062 &
  ns &
  ns &
  ns &
  0.007 \\
 &
  PAES &
  ns &
  ns &
  ns &
  0.023 &
  ns &
  ns &
  ns &
  0.016 &
  ns &
  ns &
  ns &
  0.06 \\
 &
  TDNN &
  ns &
  0.021 &
  ns &
  0.099 &
  ns &
  ns &
  ns &
  0.12 &
  ns &
  ns &
  ns &
  0.113 \\ \midrule
\textbf{} &
  \textbf{} &
  \multicolumn{4}{c|}{\textbf{Prompt 10}} &
  \multicolumn{4}{c|}{\textbf{Prompt 11}} &
  \multicolumn{4}{c}{\textbf{Prompt 12}} \\ \midrule
\multicolumn{1}{c|}{\multirow{5}{*}{Prompt-Specific}} &
  SVM (Full) &
  ns &
  ns &
  ns &
  -0.013 &
  ns &
  ns &
  ns &
  0.037 &
  ns &
  ns &
  ns &
  0.032 \\
\multicolumn{1}{c|}{} &
  SKIPFLOW-LSTM &
  ns &
  ns &
  ns &
  0.06 &
  0.017 &
  0.018 &
  ns &
  0.16 &
  ns &
  ns &
  ns &
  0.094 \\
\multicolumn{1}{c|}{} &
  CNN-LSTM-ATT &
  ns &
  ns &
  ns &
  -0.003 &
  ns &
  ns &
  ns &
  0.034 &
  ns &
  ns &
  ns &
  0.019 \\
\multicolumn{1}{c|}{} &
  R$^2$BERT &
  ns &
  ns &
  ns &
  0.018 &
  ns &
  ns &
  ns &
  0.042 &
  ns &
  ns &
  ns &
  0.018 \\
\multicolumn{1}{c|}{} &
  BERT (3 Layers) &
  ns &
  ns &
  ns &
  0.001 &
  ns &
  ns &
  ns &
  0.035 &
  ns &
  ns &
  ns &
  0.015 \\ \midrule
\multirow{4}{*}{Cross-Prompt} &
  SVM (Reduced) &
  ns &
  ns &
  ns &
  0.003 &
  ns &
  ns &
  ns &
  0.014 &
  ns &
  ns &
  ns &
  0.022 \\
 &
  Rank-SVM &
  ns &
  ns &
  ns &
  0.0 &
  ns &
  ns &
  ns &
  -0.087 &
  ns &
  ns &
  ns &
  -0.007 \\
 &
  PAES &
  ns &
  ns &
  ns &
  -0.073 &
  ns &
  ns &
  ns &
  0.056 &
  ns &
  ns &
  ns &
  0.021 \\
 &
  TDNN &
  ns &
  ns &
  ns &
  0.021 &
  ns &
  0.03 &
  ns &
  -0.065 &
  ns &
  ns &
  ns &
  0.007 \\ \bottomrule
\end{tabular}
}
\end{table*}

\begin{table*}[!htb]
 \caption{The overall \textbf{fairness} performance of the nine selected AES methods for \textbf{Race}. Cells in OSA, OSD, and CSD denote the number of prompts in which an AES method was diagnosed to have predictive bias, e.g., the number of cells with values other than `ns'. Cells in  MAED represent the average MAED of all prompts.} 
 \begin{center}
 \resizebox{0.45\textwidth}{!}{
\begin{tabular}{@{}l|l|cccc@{}}
\toprule
Types of Models & Methods & \multicolumn{1}{l}{OSA} & \multicolumn{1}{l}{OSD} & \multicolumn{1}{l}{CSD} & \multicolumn{1}{l}{MAED} \\ \midrule
\multirow{5}{*}{Prompt-Specific} & SVM (Full)       & 0 & 1 & 0 & 0.009  \\ \cmidrule(l){2-6} 
                                 & SKIPFLOW-LSTM               & 6 & 6 & 0 & 0.197  \\ \cmidrule(l){2-6} 
                                 & CNN-LSTM-ATT                & 0 & 0 & 0 & 0.005  \\ \cmidrule(l){2-6} 
                                 & R$^2$BERT & 0 & 0 & 0 & 0.019  \\ \cmidrule(l){2-6} 
                                 & BERT (3 Layers)             & 0 & 0 & 0 & 0.008  \\ \midrule
\multirow{4}{*}{Cross-Prompt}    & SVM (Reduced)      & 0 & 0 & 0 & -0.001 \\ \cmidrule(l){2-6} 
                                 & Rank-SVM                    & 1 & 3 & 1 & -0.046 \\ \cmidrule(l){2-6} 
                                 & PAES                        & 0 & 0 & 1 & -0.012 \\ \cmidrule(l){2-6} 
                                 & TDNN                        & 0 & 4 & 0 & 0.002  \\ \bottomrule
\end{tabular}
}
 \end{center}
\end{table*}

\end{document}